\documentclass[11pt]{article}

\usepackage[final]{acl}
\usepackage{times}
\usepackage{latexsym}

\usepackage[T1]{fontenc}

\usepackage[utf8]{inputenc}

\usepackage{microtype}

\usepackage{inconsolata}

\usepackage{graphicx}
\usepackage{float}
\usepackage{booktabs}
\usepackage{tabularx}
\usepackage{array}
\usepackage{makecell}

\usepackage{amsmath}
\usepackage{amssymb}

\usepackage{xspace}
\usepackage{xurl}
\usepackage{placeins}

\usepackage{algorithm}
\usepackage{algpseudocode}

\usepackage{enumitem}
\usepackage{ragged2e}

\usepackage[table]{xcolor}

\usepackage{listings}
\usepackage[most]{tcolorbox}

\newcommand{\LLM}{\mathrm{LLM}}

\newcommand{\algpar}[1]{\parbox[t]{0.86\linewidth}{\raggedright #1}}
\newcommand{\algstep}[1]{\Statex\hspace*{-0.5em}\emph{#1}}
\newcolumntype{C}{>{\centering\arraybackslash}X}

\setlist[itemize]{leftmargin=1em}
\setlist[enumerate]{leftmargin=1em}
\allowdisplaybreaks

\newcolumntype{Y}{>{\raggedright\arraybackslash}X}
\newcolumntype{L}[1]{>{\raggedright\arraybackslash}p{#1}}

\definecolor{codebg}{RGB}{248,249,251}
\definecolor{codeframe}{RGB}{218,223,230}
\definecolor{codetext}{RGB}{35,35,35}
\definecolor{linenogray}{RGB}{130,130,130}

\definecolor{kwblue}{RGB}{33,87,146}
\definecolor{typeorange}{RGB}{176,96,26}
\definecolor{funcgreen}{RGB}{24,115,84}
\definecolor{commentgray}{RGB}{120,120,120}
\definecolor{stringpurple}{RGB}{126,87,194}

\definecolor{casebg}{RGB}{248,249,251}
\definecolor{caseframe}{RGB}{205,213,224}
\definecolor{promptbg}{RGB}{252,250,244}
\definecolor{promptframe}{RGB}{222,211,184}
\definecolor{caseblue}{RGB}{33,92,168}
\definecolor{casegreen}{RGB}{0,110,120}
\definecolor{casetext}{RGB}{35,35,35}

\lstdefinestyle{cudaCase}{
  language=C++,
  basicstyle=\ttfamily\scriptsize\color{codetext},
  numbers=left,
  numberstyle=\tiny\color{linenogray},
  numbersep=6pt,
  frame=single,
  framerule=0.6pt,
  rulecolor=\color{codeframe},
  backgroundcolor=\color{codebg},
  xleftmargin=1.6em,
  framexleftmargin=1.2em,
  columns=fullflexible,
  keepspaces=true,
  breaklines=false,
  showstringspaces=false,
  tabsize=2,
  aboveskip=0.5em,
  belowskip=0.5em,
  keywordstyle=\color{kwblue}\bfseries,
  commentstyle=\color{commentgray}\itshape,
  stringstyle=\color{stringpurple},
  morekeywords={
    __global__,__device__,__forceinline__,return,for,const,int,void
  },
  morekeywords=[2]{
    float,float4
  },
  keywordstyle=[2]\color{typeorange}\bfseries,
  emph={
    _swish_fast,__ldg,reinterpret_cast,
    swish_vec4_ldg_kernel,swish_vec4_ldg_forward,
    torch,Tensor
  },
  emphstyle=\color{funcgreen}\bfseries
}

\lstdefinestyle{swishstyle}{
  language=C++,
  basicstyle=\ttfamily\scriptsize\color{codetext},
  keywordstyle=\bfseries\color{kwblue},
  commentstyle=\itshape\color{commentgray},
  showstringspaces=false,
  keepspaces=true,
  breaklines=false,
  columns=fullflexible,
  aboveskip=0pt,
  belowskip=0pt,
  xleftmargin=0pt,
  xrightmargin=0pt,
  emph={float4,__ldg,_swish_fast,stride,n4},
  emphstyle=\bfseries\color{funcgreen}
}

\lstdefinestyle{casepromptstyle}{
  basicstyle=\ttfamily\scriptsize\color{casetext},
  breaklines=true,
  breakatwhitespace=false,
  columns=fullflexible,
  keepspaces=true,
  showstringspaces=false,
  aboveskip=0pt,
  belowskip=0pt
}

\lstdefinestyle{casecodestyle}{
  language=Python,
  basicstyle=\ttfamily\scriptsize\color{casetext},
  keywordstyle=\bfseries\color{caseblue},
  commentstyle=\itshape\color{commentgray},
  stringstyle=\color{casegreen},
  breaklines=true,
  breakatwhitespace=false,
  columns=fullflexible,
  keepspaces=true,
  showstringspaces=false,
  aboveskip=0pt,
  belowskip=0pt,
  emph={float4,__ldg,_swish_fast,load_inline,ModelNew,torch},
  emphstyle=\bfseries\color{casegreen}
}

\newtcolorbox{casecard}[1]{
  enhanced,
  breakable,
  colback=casebg,
  colframe=caseframe,
  boxrule=0.45pt,
  arc=1.5pt,
  left=6pt,
  right=6pt,
  top=5pt,
  bottom=5pt,
  title={#1},
  fonttitle=\bfseries\small,
  colbacktitle=casebg,
  coltitle=black
}

\newtcblisting{casepromptbox}[1]{
  enhanced,
  breakable,
  listing only,
  colback=promptbg,
  colframe=promptframe,
  boxrule=0.45pt,
  arc=1.5pt,
  left=5pt,
  right=5pt,
  top=5pt,
  bottom=5pt,
  title={#1},
  fonttitle=\bfseries\small,
  colbacktitle=promptbg,
  coltitle=black,
  listing options={style=casepromptstyle}
}
\tcbset{
  casecodebase/.style={
    enhanced,
    breakable,
    listing only,
    colback=codebg,
    colframe=codeframe,
    boxrule=0.45pt,
    arc=1.5pt,
    left=5pt,
    right=5pt,
    top=5pt,
    bottom=5pt,
    fonttitle=\bfseries\small,
    colbacktitle=codebg,
    coltitle=black,
  }
}
\newtcblisting{casecodebox}[2][]{
  casecodebase,
  title={#2},
  listing options={style=casecodestyle},
  #1
}

\title{HTAM: Hierarchical Transition-Attended Memory for Operator Optimization}

\author{First Author \\
  Affiliation / Address line 1 \\
  Affiliation / Address line 2 \\
  Affiliation / Address line 3 \\
  \texttt{email@domain} \\\And
  Second Author \\
  Affiliation / Address line 1 \\
  Affiliation / Address line 2 \\
  Affiliation / Address line 3 \\
  \texttt{email@domain} \\}

\author{
\textbf{Yining Zhang}$^{1,2,3}$ \quad
\textbf{Mingyang Yi}$^{3,4,*}$ \quad
\textbf{Chen Wang}$^{3}$ \quad
\textbf{Xuwen Xiang}$^{3}$ \\
\textbf{Tianhe Jia}$^{3}$ \quad
\textbf{Zedong Dan}$^{3}$ \quad
\textbf{Chengqing Zong}$^{1,2}$ \quad
\textbf{Yue Wang}$^{3,5,*}$ \\
\\[-0.5em]
\begin{tabular}{c}
$^{1}$School of Artificial Intelligence, University of Chinese Academy of Sciences \\
$^{2}$Institute of Automation, Chinese Academy of Sciences \\
$^{3}$Zhongguancun Academy \quad
$^{4}$Renmin University of China \quad
$^{5}$XinZhuAI  \\
\texttt{zhangyining2024@ia.ac.cn, yimingyang@ruc.edu.cn, yuewang@bza.edu.cn}
\end{tabular}
}

\begin{document}
\maketitle

\begingroup
\renewcommand{\thefootnote}{*}
\footnotetext{Corresponding authors.}
\endgroup

\begin{abstract}

High-performance GPU kernels are essential for efficient LLM deployment, yet optimizing them remains expertise-intensive. Recent LLM-based code generation makes automatic GPU operator generation promising, but operator optimization remains a hardware-aware search problem.
Existing LLM-based methods face a granularity mismatch: coarse hints are reusable but hard to execute, whereas detailed memories are actionable but enlarge the search space and obscure optimization bottlenecks. The key challenge is therefore to organize optimization experience at an appropriate granularity.
To address this issue, this paper proposes \textbf{HTAM} (\textbf{H}ierarchical \textbf{T}ransition-\textbf{A}ttended \textbf{M}emory), a coarse-to-fine framework for LLM-based operator optimization.
HTAM builds a two-level Hierarchical Transition Graph (HTG) to organize coarse global directions, detailed local strategies, and transition experience between optimization steps.
During each evolution step, HTAM selects a global direction from the current state and recent optimization history, retrieves the corresponding local strategy memory, and uses it to guide concrete CUDA code generation.
Experiments on KernelBench demonstrate that HTAM improves correctness, fast-solution rate, and speedup over LLM-based baselines, while backend and Robust-KBench studies indicate transferable benefits from structured memory.

\end{abstract}

\section{Introduction}
\label{sec:intro}
The rapid scaling of large language models (LLMs), together with the continuous evolution of vendor-specific accelerator architectures such as GPUs, has created an increasing demand for high-performance operator development~\citep{miao2025towards, ye2025flashinfer, tillet2019triton, wang2025tilelang}.
Writing efficient GPU kernels, however, remains a labor-intensive process that requires substantial hardware expertise and engineering effort~\citep{ouyang2025kernelbench, naveed2025comprehensive, faingnaert2021flexible}. 
Recent LLMs have demonstrated strong code generation capabilities~\citep{achiam2023gpt,chen2021evaluating,roziere2023code,guo2024deepseek,liu2025deepseek}, making it promising to use LLMs to automatically generate high-performance kernels~\citep{chen2025cuda,han2026making}.

\begin{figure}[t]
    \centering
    \includegraphics[width=\columnwidth]{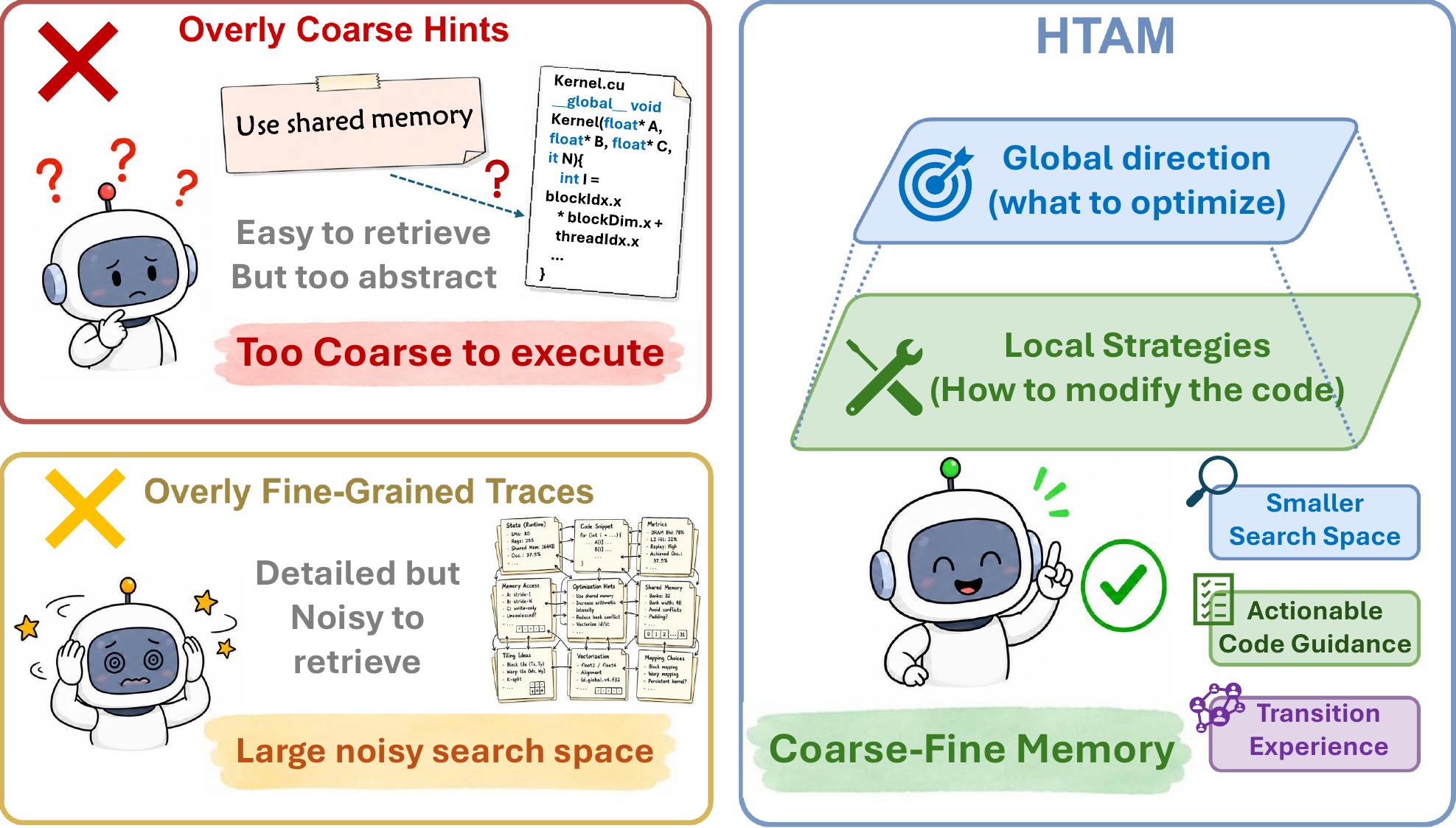}
    \caption{HTAM completes operator optimization in a coarse-to-fine manner, organizes optimization experience into high-level directions, concrete strategies, and transition relations.}
    \label{fig:htam_teaser_single}
\end{figure}

However, developing high-performance GPU operators is highly hardware-aware.
The complexity comes from the multi-level GPU memory hierarchy, increasingly specialized hardware units, and architecture-specific execution constraints.
Searching for an efficient kernel implementation in such a large design space often requires iterative or evolutionary refinement rather than one-shot generation~\citep{novikov2025alphaevolve,dai2026cuda,lange2025towards}.
This poses two key challenges.
(1) Kernel optimization requires coupled multi-stage decisions: an effective kernel must jointly choose and combine tile sizes, thread/block mapping, memory access patterns, data reuse, and synchronization strategies.
(2) Implementing a selected optimization strategy is non-trivial: it must still be translated into correct hardware-aware CUDA code.
Together, these challenges make optimizing memory difficult to organize: it must guide a multi-step search while still providing concrete evidence for executable code edits.

Existing methods organize optimization experience at either overly coarse or overly fine granularity (Fig.~\ref{fig:htam_teaser_single}). 
CudaForge uses coarse-grained bottleneck diagnoses and textual hints, which are easy to retrieve but too abstract for executable CUDA code generation~\citep{zhang2025cudaforge}. 
Robust-KBench instead relies on fine-grained implementation traces, which provide richer details but create a large and noisy state--strategy search space, making retrieval and reuse inefficient~\citep{lange2025towards}.

To address this granularity mismatch, this paper proposes \textbf{HTAM}
(\textbf{H}ierarchical \textbf{T}ransition-\textbf{A}ttended \textbf{M}emory),
a hierarchical memory framework for LLM-based operator optimization.
HTAM organizes and leverages historical
optimization experience in a coarse-to-fine manner, mirroring the decision
process of human experts: it first retrieves high-level experience to select a
global optimization direction, and then uses strategy-level experience to guide
concrete CUDA optimization implementation.

HTAM realizes this process with a \textbf{Hierarchical Transition Graph} (HTG), where global nodes store direction-level experience and associated local nodes store implementation-level strategy.
To model dependencies across optimization steps, HTG stores transition experience on edges between global nodes and aggregates recent history with an attention-inspired rule~\citep{vaswani2017attention}.
This replaces costly sequence-level memory with reusable node-to-node transition experience, reducing the complexity and pre-exploration cost of the memory bank. Our contributions are as follows: 
\begin{itemize}
    \item HTAM introduces a coarse-to-fine optimization framework that decomposes LLM-based operator optimization into global direction selection and local strategy refinement.

    \item HTAM organizes optimization experience into hierarchical node and edge memories and introduces transition-aware scoring to guide multi-step decisions, while reducing memory storage and pre-exploration costs.
    
    \item Extensive experiments show that HTAM substantially improves over controlled reproduced baselines on the full KernelBench suite, reaching \textbf{98.4\%} correctness, \textbf{84.0\%} Fast@1, and \textbf{1.978$\times$} geometric-mean speedup, while backend and Robust-KBench studies provide preliminary evidence of generalization.
\end{itemize}

\section{Related Work}

\paragraph{LLM-based Kernel and Operator Optimization.}
Recent LLM-based methods generate and optimize CUDA kernels with code generation, test-time scaling, and execution feedback such as compilation, correctness, runtime, and profiling signals~\citep{chen2025cuda,han2026making,jin2025reveal,li2026ets,muennighoff2025s1,bagirov2025best,zhang2025cudaforge}; agentic and hierarchical systems further decompose optimization into planning and implementation stages~\citep{wei2025astra,zhu2025qimeng}.
However, accumulated experience remains weakly structured for reuse across directions, strategies, and transitions.

\paragraph{Memory and Graph-based Evolution.}
The prior domain knowledge and reusable past interactions have also been proven to be useful for LLM generation~\citep{li2026reasoning,lewis2020retrieval,zhong2024memorybank,packer2023memgpt,chhikara2025mem0,xu2025mem,wang2026tandem,novikov2025alphaevolve}. 
For example, in CUDA optimization, KernelBlaster \citep{dong2026kernelblaster} maintains a persistent CUDA knowledge base, and From Large to Small~\citep{gong2025large} transfers CUDA optimization expertise through a reasoning graph. 
While these works show the value of reusable experience, their memory organizations can still lead to large search spaces and inefficient reuse.

\section{Problem Setup}\label{sec:problem setup}

\begin{figure}[t]
    \centering
    \includegraphics[width=\columnwidth]{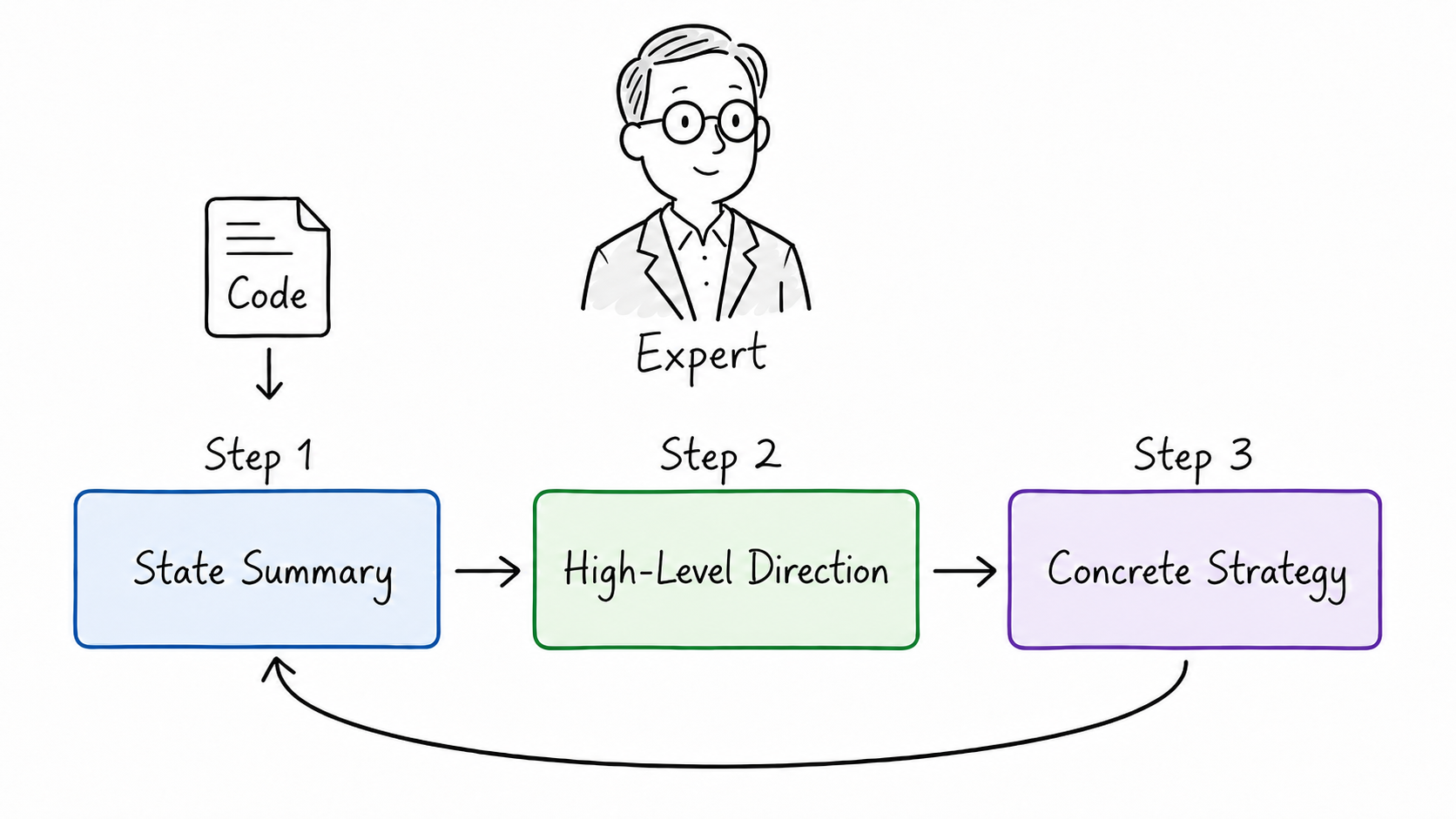}
    \caption{Human expert operator optimization typically follows three steps: state summarization, high-level direction selection, and concrete strategy selection.}
    \label{fig:human_expert_workflow}
\end{figure}

As illustrated in Figure~\ref{fig:human_expert_workflow}, human experts typically optimize an operator iteratively: First, summarize the current implementation state, then determine a high-level optimization direction, select a concrete optimization strategy, and continue refining the implementation with feedback.~\citep{chen2018tvm,zheng2023ansorgenerating,zhu2022roller,dao2023flashattention2}

Inspired by this expert workflow, we model operator optimization as a $T$-step evolution process, where $c_t$ denotes the operator implementation at step $t$. At each step, we maintain a state $s_t$ that summarizes useful historical information up to step $t$; the complete state fields are provided in Appendix~\ref{app:state_fields}. Given $c_t$ and an evolution strategy $a_t$, the next implementation is generated as

\begin{equation}
c_{t + 1} = \LLM(c_{t}, s_{t}, a_{t}, \mathcal{M}),  
\end{equation}
by providing the current implementation $c_{t}$, state $s_{t}$ (a summarization of related information), experience from $\mathcal{M}$, and evolution strategy $a_{t}$ to LLM. Here the strategy $a_{t}$ is decided by a policy 
\begin{equation}\label{eq:action policy}
a_{t}\sim \pi(\cdot \mid c_{t}, s_{t}, \mathcal{M}),
\end{equation}
which will be specified in Section~\ref{sec:how_to_use}. 

During the evolution process, the state and memory bank will be progressively updated as 
\begin{equation}
s_{t + 1} = \LLM(s_{t}, c_{t + 1}, a_{ t}),
\end{equation}
by summarizing state $s_{t}$, current implementation $c_{t + 1}$ and action $a_{t}$ with LLM, and 
\begin{equation}
\mathcal{M} = \mathcal{M}\bigcup \{s_{t}, c_{t + 1}, a_{t}\}.
\end{equation}

The objective is to progressively move the optimization process toward a correct and faster implementation within finite evolution steps:

\begin{equation}
\min_{\substack{0 \leq t \leq T \\
                c_t \ \mathrm{is\  correct}}}
\mathrm{runtime}(c_t).
\end{equation}

Since each implementation depends on previous optimization decisions and feedback, operator optimization is naturally a sequential decision problem.


\section{Hierarchical Transition-Attended Memory (HTAM)}

In this section, we present \textbf{HTAM} 
(\textbf{H}ierarchical \textbf{T}ransition-\textbf{A}ttended \textbf{M}emory), 
a hierarchical framework for LLM-based operator optimization.
We describe HTAM through three questions: how to organize memory, what to store, and how to use the memory to guide operator optimization.

\begin{figure*}[t]
    \centering
    \includegraphics[width=0.96\textwidth]{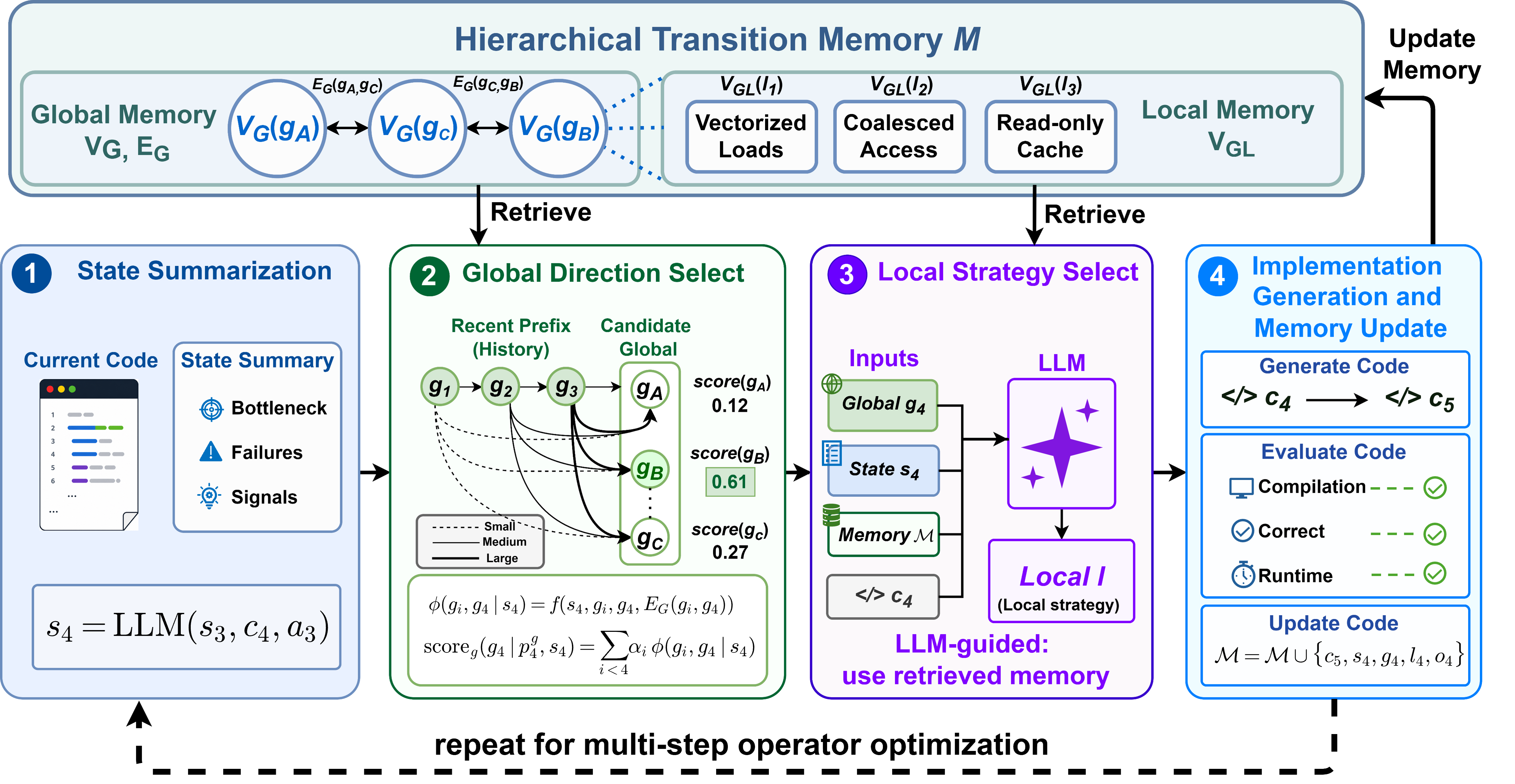}
    \caption{Overall framework of HTAM. The current implementation and evaluation feedback are summarized into a compact state, which is then used to retrieve and rank candidate directions and strategies from a hierarchical graph memory. After action execution, the resulting feedback is written back to the node and edge memories.}
    \label{fig:htam_framework}
\end{figure*}

\subsection{How to Organize the Memory Bank: Hierarchical Transition Graph}
When human experts conduct operator optimization, they often follow a ``global-direction-first, local-strategy-second'' process. To mirror this process, our HTAM organizes the memory bank as a \emph{Hierarchical Transition Graph (HTG)}
\begin{equation}\label{eq:HTG}
\mathcal{M} = (V_G, V_{GL}, E_{G}),
\end{equation}
where $V_G$ denotes global nodes, $V_{GL}$ are the specific local nodes under each global node, and $E_{G}$ denotes transition edges between global optimization directions.

\textbf{Global nodes $V_{G}$} encode coarse optimization directions that generalize across operators.
Each global direction $g_i$ corresponds to a node $V_G(g_i)$, which captures a high-level intent guiding implementation evolution. Examples include memory access, data reuse, instruction throughput, boundary handling, and parallelism optimization. The full list is provided in Appendix~\ref{app:global-nodes}.

\textbf{Local nodes $V_{GL}$} encode executable strategies under parent global nodes, with each strategy $l_i$ mapped to a local node $V_{GL}(l_i)$.
For example, under \textit{Memory Access}, strategies include aligned vectorized loads and read-only-load promotion.
The details of local nodes are in Appendix~\ref{app:local-nodes}.

\textbf{Global transition edges $E_{G}$} connect global nodes, e.g. $E_{G}(g_{i}, g_{j})$ is the edge between nodes $g_{i}$ and $g_{j}$. It will be further leveraged to guide the transition from one global node to another.  

The designed HTG avoids selections over a large number of local strategies and makes the search process more structured. It is also aligned with human experts' memory bank and recent efforts on organized and dynamically managed memory for LLM systems \citep{packer2023memgpt,xu2025mem}.

\begin{figure}[t]
    \centering
    \includegraphics[width=\columnwidth]{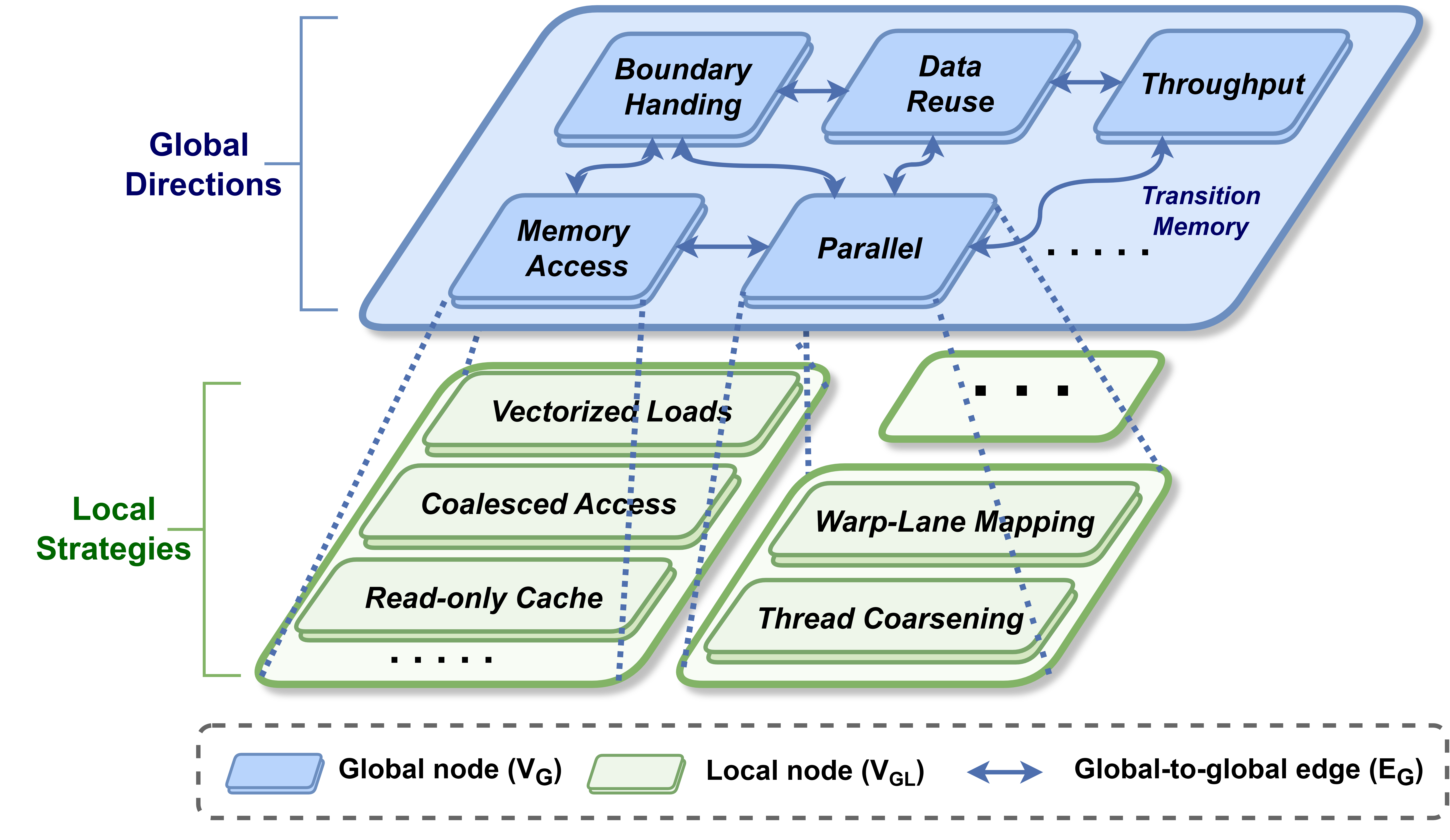}
    \caption{Hierarchical Transition Graph (HTG), where $V_G$ contains global directions, $V_{GL}$ contains local strategies, and $E_G$ stores global-to-global transitions.}
    \label{fig:htam_graph}
\end{figure}

\subsection{What to Store in HTG: Structured Node and Edge Memories}

After defining the HTG structure, we specify its memory contents.
As shown in Figure~\ref{fig:htam_graph}, HTG provides a compact CUDA-oriented schema, where nodes and transition edges are readable, updatable memories combining static priors, runtime feedback, and examples.

\textbf{Global nodes} store high-level optimization knowledge.
Each node represents a coarse direction with static priors, including optimization goals, triggers, applicable patterns, risks, and expected gain types.
It also maintains runtime evidence such as attempt frequency, success statistics, and improvement trends.
Together, these fields define the intent and indicate when it is empirically useful.

\textbf{Local nodes} store implementation-level strategy knowledge.
Each node acts as an expert-inspired strategy card under a parent global direction, specifying usage conditions, edit patterns, verification checks, and expected effects.
Since local strategies are instantiated as code edits, each local node maintains positive and negative evidence explaining why the strategy succeeded or failed in similar contexts, helping the LLM optimize code implementation more effectively.

\textbf{Global transition edges} store path-dependent knowledge between global intents.
Each edge represents a reusable transition between two global directions, recording when the transition is applicable, why it may help, and what risks it may introduce.
It also stores trajectory-level evidence for this transition, including performance gains, compilation/correctness outcomes, and downstream returns.
Detailed fields are provided in Appendix~\ref{app:htg-memory-contents}.
The next subsection explains how these memories are read and updated for action selection.

\subsection{How to Use Memory: Transition-Aware Action Selection and Memory Update}
\label{sec:how_to_use}

With these organized memories, HTAM guides kernel optimization through an iterative evolution process. 
At each step, it selects a global direction from the current state and relevant memories, then decides a local strategy to guide the code generation.
After executing the evolved code, HTAM records evaluation feedback, updates the memory bank, and proceeds to the next evolution step.

\textbf{State Summarization.}
Following Section~\ref{sec:problem setup}, HTAM summarizes the current code and history into a compact state $s_t$, including task/code context, compile/correct/runtime feedback, best-so-far speedup, stagnation signals, recent global/local choices, and code symptoms.

\textbf{Global Direction Selection.}  With the summarized state, we first decide the global optimization direction. 
Let the global optimization prefix (directions) before step $t$ be
\begin{equation}\label{eq:prefix_direction}
p_t^g=(g_1,\ldots,g_{t-1}), \quad g_i\in V_{G}.
\end{equation}

For the first step, HTAM selects $g_1$ from node-level HTG statistics without using transition memory. For each candidate global direction $g_t\in V_G$, HTAM reads the edge memories $E_G(g_i,g_t)$ connecting it to previous global directions $g_i$. We then define the ``global-to-global'' transition utility under the current state $s_t$ as
\begin{equation}
\phi(g_i, g_t \mid s_t)
=
f\!\left(s_t, g_i, g_t, E_G(g_i,g_t)\right),
\end{equation}
where $f(\cdot)$ is a deterministic feature-based scorer rather than an LLM-inferred score; it maps state-matched transition evidence on $E_G(g_i,g_t)$ into a scalar utility. Details are in Appendix~\ref{app:transition_reward}.

Then, to evaluate the transition from the current state $s_t$ to a global direction $g_t$, we define:
\begin{equation}\label{eq:score}
\operatorname{score}_{g}(g_{t}\mid p_t^g, s_t)
=
\sum_{i<t}\alpha_i\,\phi(g_i,g_t\mid s_t),
\end{equation}
where $\alpha_i$ is a position-aware weight inspired by attention-style weighting~\citep{vaswani2017attention}:
\begin{equation}\label{eq:prefix_weight}
\alpha_i = \exp\left(-\lambda(|p_t^g|-i)\right),
\quad i=1,\ldots,|p_t^g|.
\end{equation}
Notably, our score \eqref{eq:score} relates the candidate direction $g_t$ to the prefix $p_t^g$ through edge memories $\{E_{g_i,g_t}\}_{i<t}$.
A direct alternative is to store full-prefix transition experience $p_t^g\!\to g_t$, but its space grows as $|V_G|^t$ and is impractical to cover.
HTAM instead decomposes prefix experience into reusable pairwise edge memories, significantly reducing the cost of building the memory bank (from $|V_{G}|^{T}\to |V_{G}|^{2}$).

With the scores on global direction in $\mathcal{G}$, we define a Softmax-based policy on global direction: 
\begin{equation}\label{eq:global direction}
\begin{aligned}
& \mu_g(g_{t}\mid p_t^g,s_t) 
 = \epsilon\mathrm{Unif}_{\mathcal{G}_t} \\
 & + (1 - \epsilon)\frac{\exp\left(\operatorname{score}_g(g_{t}\mid p_t^g,s_t) / \tau\right)}{\sum_{g\in\mathcal{G}_t}\exp\left(\operatorname{score}_g(g\mid p_t^g,s_t) / \tau\right)}, 
\end{aligned} 
\end{equation}
where the first term provides ``$\epsilon$-greedy'' exploration, and the second term selects a global direction with a higher score.
In this way, global selection exploits high-reward transition patterns while preserving exploration over less-tried directions.

\textbf{Local Strategy Selection.} After selecting the global direction with \eqref{eq:global direction}, HTAM restricts local strategy candidates to the associated local nodes:
\begin{equation}\label{eq:local nodes}
\mathcal{L}(g_t)=\{l\in V_{GL}:\operatorname{par}(l)=V_{G}(g_t)\},
\end{equation}
where $\operatorname{par}(l)$ denotes parent global node of $l$.
Local strategy selection conditions on state $s_t$, implementation $c_t$, global-node memory $V_G(g_t)$, and local-node memory $V_{GL}(l)$:
\begin{equation}\label{eq:local policy}
l_t
=
\arg\max_{l\in\mathcal{L}(g_t)}
\mathrm{LLM}\left(s_t,c_t, V_{G}(g_{t}), V_{GL}(l)\right).
\end{equation}

Here, $\mathrm{LLM}(s_t,c_t, V_G(g_t), V_{GL}(l))$ summarizes the information from the current state, implementation, selected global node, and candidate local node into a scalar preference score. The prompt template and output schema are provided in Appendix~\ref{app:local_selection_prompt}.

\textbf{Implementation Generation and Memory Update.} Given the local strategy $l_t$, the LLM generates the next implementation as
\begin{equation}\label{eq:generation}
    c_{t + 1} = \mathrm{LLM}(c_{t}, s_{t}, l_{t}, V_{GL}(l_{t})),
\end{equation}
where $V_{GL}(l_t)$ provides retrieved local examples relevant to $s_t$ and $c_t$.
To mirror the notations in Section~\ref{sec:problem setup}, we clarify that the evolution action $a_{t}$ is the composition of global direction $g_{t}$ and local strategy $l_{t}$, and the action policy \eqref{eq:action policy} is the composition of (global/local) policies \eqref{eq:global direction} and \eqref{eq:local policy}. 

\begin{algorithm}[t]
\footnotesize
\setlength{\baselineskip}{9pt}
\setlength{\itemsep}{0pt}
\caption{HTAM Optimization Procedure}
\label{alg:htam}
\begin{algorithmic}[1]
\Require Initial implementation $c_0$, initialized HTG memory bank $\mathcal{M}_0=(V_G,V_{GL},E_G)$, evolution steps $T$
\Ensure Best candidate $c^\star$

\State $c_1,c^\star\gets c_0;\ \mathcal{M}\gets\mathcal{M}_0;\ s_0,a_0\gets\emptyset;\ p_1^g\gets()$

\For{$t = 1, \ldots, T$}
    \algstep{Step 1. State Summarization.}
    \State $s_t \gets \mathrm{LLM}(s_{t-1}, c_t, a_{t-1})$

    \algstep{Step 2. Global Direction Selection.}
    \ForAll{$g \in V_G$}
        \State Compute $\mathrm{score}_g(g \mid p_t^g, s_t)$ by \eqref{eq:score}
    \EndFor
    \State Select global direction $g_t \sim \mu_g(\cdot \mid p_t^g, s_t)$ \eqref{eq:global direction}

    \algstep{Step 3. Local Strategy Selection.}
    \State Aggregate local strategies $\mathcal{L}_t$ in \eqref{eq:local nodes}
    \ForAll{$l \in \mathcal{L}_t$}
        \State \algpar{Evaluate $l$ by $\mathrm{LLM}(s_t, c_t, V_G(g_t), V_{GL}(l))$}
    \EndFor
    \State Select the best-scoring strategy $l_t$

    \algstep{Step 4. Implementation Generation and Memory Update.}
    \State \algpar{Generate $c_{t+1} \gets \mathrm{LLM}(c_t, s_t, l_t, V_{GL}(l_t))$}

    \State Evaluate $c_{t+1}$ to get feedback $o_t$

    \If{$c_{t+1}$ is better than $c^\star$}
        \State $c^\star \gets c_{t+1}$
    \EndIf
    \State \algpar{Update $\mathcal{M} \gets \mathcal{M} \cup \{c_{t+1}, s_t, g_t, l_t, o_t\}$}
    \State $p_{t+1}^g \gets \Call{Append}{p_t^g, g_t}$
\EndFor

\State \Return $c^\star$
\end{algorithmic}
\end{algorithm}

\FloatBarrier

Finally, after evaluating the evolved implementation $c_{t+1}$, HTAM updates the resulting feedback $o_{t}$ stored in memory bank $\mathcal{M}$:
\begin{equation}\label{eq:memory_update}
\mathcal{M} = \mathcal{M}\bigcup \{c_{t + 1}, s_{t}, g_{t}, l_{t}, o_{t}\}.
\end{equation}
Concretely, this refreshes the global-node statistics, local-node evidence, and transition-edge feedback for $V_G(g_t)$, $V_{GL}(l_t)$, and $E_G(g_{t-1},g_t)$ when $t>1$.

\begin{table*}[t]
\centering
\scriptsize
\setlength{\tabcolsep}{2.5pt}
\renewcommand{\arraystretch}{1.10}
\begin{tabular*}{\textwidth}{@{\extracolsep{\fill}}rlcccccccccc}
\toprule
& & \multicolumn{4}{c}{\textbf{Strategy}} 
& \multicolumn{3}{c}{\textbf{Overall Performance}} 
& \multicolumn{3}{c}{\textbf{KernelBench Spd.}} \\
\cmidrule(lr){3-6} \cmidrule(lr){7-9} \cmidrule(lr){10-12}
\# 
& Method 
& TTS 
& Evo 
& Mem
& HieDec
& Correct$\uparrow$ 
& Fast@1$\uparrow$ 
& GeoM Spd.$\uparrow$ 
& L1$\uparrow$ 
& L2$\uparrow$ 
& L3$\uparrow$ \\
\midrule

\rowcolor{gray!12}
\multicolumn{12}{c}{\textbf{Vanilla LLM Generation}} \\
\midrule
1 & DeepSeek-R1 & $\times$ & $\times$ & $\times$ & $\times$ & 74.4\% & 14.8\% & 0.827$\times$ & 0.705$\times$ & 0.926$\times$ & 0.907$\times$ \\
2 & DeepSeek-V3 & $\times$ & $\times$ & $\times$ & $\times$ & 82.0\% & 7.2\% & 0.194$\times$ & 0.060$\times$ & 0.314$\times$ & 0.780$\times$ \\
3 & DeepSeek-V4-Flash & $\times$ & $\times$ & $\times$ & $\times$ & 64.0\% & 10.0\% & 0.330$\times$ & 0.171$\times$ & 0.375$\times$ & 0.952$\times$ \\
4 & Gemini-2.5-Flash-Thinking & $\times$ & $\times$ & $\times$ & $\times$ & 86.0\% & 22.8\% & 0.355$\times$ & 0.099$\times$ & 0.904$\times$ & 0.699$\times$ \\
5 & OpenAI-o3$^\dagger$ & $\times$ & $\times$ & $\times$ & $\times$ & 57.6\% & 31.6\% & 0.680$\times$ & -- & -- & -- \\

\midrule
\rowcolor{gray!12}
\multicolumn{12}{c}{\textbf{Vanilla Evolution-based Methods}} \\
\midrule
6 & Best-of-$N$ Sampling & $\checkmark$ & $\times$ & $\times$ & $\times$ & 80.4\% & 24.8\% & 0.909$\times$ & 0.780$\times$ & 1.030$\times$ & 0.960$\times$ \\
7 & Feedback-only Refinement & $\checkmark$ & $\checkmark$ & $\times$ & $\times$ & 84.8\% & 49.2\% & 1.065$\times$ & 0.955$\times$ & 1.180$\times$ & 1.080$\times$ \\
8 & Flat Memory Retrieval & $\checkmark$ & $\checkmark$ & $\checkmark$ & $\times$ & 89.6\% & 58.8\% & 1.464$\times$ & 1.194$\times$ & 1.917$\times$ & 1.284$\times$ \\

\midrule
\rowcolor{gray!12}
\multicolumn{12}{c}{\textbf{Refined Evolution-based Methods}} \\
\midrule
9 & KernelBlaster$^\ddagger$ \citep{dong2026kernelblaster} & $\checkmark$ & $\checkmark$ & $\checkmark$ & $\times$ & 80.2\% & 62.8\% & 1.756$\times$ & 1.497$\times$ & 2.592$\times$ & 1.110$\times$ \\
10 & CudaForge$^\ddagger$ \citep{zhang2025cudaforge} & $\checkmark$ & $\checkmark$ & $\times$ & $\times$ & 97.6\% & 70.8\% & 1.677$\times$ & 1.448$\times$ & 2.104$\times$ & 1.283$\times$ \\

\midrule
\rowcolor{gray!12}
\multicolumn{12}{c}{\textbf{Ours}} \\
\midrule
11 & \textbf{HTAM} & $\checkmark$ & $\checkmark$ & $\checkmark$ & $\checkmark$ & \textbf{98.4\%} & \textbf{84.0\%} & \textbf{1.978$\times$} & \textbf{1.532$\times$} & \textbf{2.598$\times$} & \textbf{1.909$\times$} \\

\bottomrule
\end{tabular*}
\caption{
Main comparisons on KernelBench. Reproduced rows follow the official KernelBench protocol; speedup is the geometric mean over tasks, measured against the PyTorch eager reference, and L1--L3 report level-wise speedups. TTS/Evo/Mem/HieDec denote test-time scaling, evolution, memory, and hierarchical decision. Rows marked with $\dagger$ or $\ddagger$ are reported external results included for context; see Appendix~\ref{app:reported_systems}.
}

\label{tab:main_kernelbench_leaderboard}
\end{table*}

Overall, HTAM forms a memory-assisted state-dependent evolution loop: in each step, transition memories guide global direction; local strategy guides the code generation, and execution feedback is converted into reusable structured memory. 
The complete procedure is summarized in Algorithm~\ref{alg:htam}.


\section{Experiments}
\label{sec:experiments}
\subsection{Experimental Setup}

We evaluate HTAM on KernelBench~\citep{ouyang2025kernelbench}, using the full 250-task suite: 100 Level-1 single-operator tasks, 100 Level-2 operator-fusion tasks, and 50 Level-3 model-level tasks. 
All evaluations are conducted on NVIDIA A100-SXM4-80GB GPUs with CUDA 12.8, PyTorch 2.9.1+cu128, and the official KernelBench v0.2.0.dev0. HTAM and reproduced baseline runs use DeepSeek-R1~\citep{guo2025deepseek} as the backbone model.
The main metrics are correctness, Fast@1, and geometric mean speedup following the official KernelBench protocol.
Detailed task coverage, hardware configuration, timing protocol, and metric definitions are provided in Appendix~\ref{app:exp_protocol}.

\subsection{Baseline Methods}
We compare HTAM with three groups of baselines; details are provided in Appendix~\ref{app:baseline_details}.

\textbf{Vanilla LLM Generation.}
We use single-pass generation with general-purpose LLMs, including DeepSeek-R1~\citep{guo2025deepseek}, DeepSeek-V3~\citep{liu2024deepseek}, DeepSeek-V4-Flash~\citep{deepseekai2026deepseekv4}, Gemini-2.5-Flash-Thinking~\citep{comanici2025gemini}, and the OpenAI-o3 result from CudaForge~\citep{zhang2025cudaforge}.

\textbf{Vanilla Evolution-based Methods.}
We include Best-of-$N$ Sampling~\citep{brown2024large}, Feedback-only Refinement~\citep{shinn2023reflexion}, and Flat Memory Retrieval~\citep{novikov2025alphaevolve}, all reproduced with the same DeepSeek-R1 backbone and evaluation protocol as HTAM.

\textbf{Refined Evolution-based Methods.}
We further compare with KernelBlaster~\citep{dong2026kernelblaster} and CudaForge~\citep{zhang2025cudaforge} as external reported references, since they may use different models, budgets, or metric definitions.

\subsection{Main Results}

Table~\ref{tab:main_kernelbench_leaderboard} presents the main comparison on KernelBench.
Three key observations emerge.

\textbf{HTAM is both reliable and effective.}
HTAM achieves the best results among controlled reproduced baselines, with $98.4\%$ correctness, $84.0\%$ Fast@1, and a $1.978\times$ geometric mean speedup. Compared with the strongest reproduced baseline, Flat Memory Retrieval, HTAM improves correctness by $8.8$ percentage points, Fast@1 by $25.2$ percentage points, and GeoM speedup from $1.464\times$ to $1.978\times$. This shows that HTAM improves optimization reliability while generating faster kernels.

\textbf{HTAM remains effective across task levels.}
HTAM obtains $1.532\times$, $2.598\times$, and $1.909\times$ geometric mean speedups on L1, L2, and L3 tasks, respectively. The larger gain on L2 is consistent with the KernelBench task structure, where operator-fusion tasks expose richer memory-access, data-reuse, and layout-related opportunities.
The consistent improvements across all three levels show that HTAM is not specialized to a single task type.

\textbf{Memory organization matters.}
Direct generation with the same base model reaches only $74.4\%$ correctness, $14.8\%$ Fast@1, and $0.827\times$ GeoM speedup, and Best-of-$N$ sampling remains limited.
Feedback-only refinement and Flat Memory Retrieval improve over direct generation, but still lag behind HTAM. This suggests that the key factor is not merely feedback, sampling, or memory reuse, but how optimization experience is organized and transferred across global directions, local strategies, and step transitions.
Appendix~\ref{app:api_cost} further shows that, with the reusable memory bank amortized, the six-step online cost of optimizing one operator is about \$0.21 under DeepSeek-R1 API pricing.

\begin{table}[t]
\centering
\small
\setlength{\tabcolsep}{3.0pt}
\renewcommand{\arraystretch}{1.08}
\begin{tabular*}{\columnwidth}{@{\extracolsep{\fill}}lcccc}
\toprule
\textbf{Code LLM} 
& \textbf{Corr.$\uparrow$}
& \textbf{Fast$\uparrow$}
& \textbf{Spd.$\uparrow$}
& \textbf{L1/L2/L3 Spd.$\uparrow$} \\
\midrule
DS-V4F
& 88.0\%
& 58.0\%
& 1.472$\times$
& 1.340 / 1.720 / 1.300 \\

DS-R1
& \textbf{98.0\%}
& 84.0\%
& 2.003$\times$
& \textbf{1.560} / 2.620 / 1.930 \\

GPT-4o
& 96.0\%
& \textbf{86.0\%}
& \textbf{2.037$\times$}
& 1.520 / \textbf{2.720} / \textbf{2.050} \\
\bottomrule
\end{tabular*}
\caption{
Generalization across LLM backends on a 50-task KernelBench subset. Correct and Fast@1 are percentages; geometric-mean speedups (Spd.) are in $\times$.
}
\label{tab:llm_backend}
\end{table}

\subsection{Backend and Transfer Analysis}
\label{sec:generalization}
We further study whether HTAM's structured memory remains transferable after organization, both across different LLM backends and beyond the original KernelBench task suite.

\paragraph{LLM Backends Generalization.}
\label{sec:llm_backend}

We explore whether HTAM depends on a specific code-generation API by evaluating a stratified 50-task KernelBench subset, since running multiple LLM backends under the full search budget is expensive.
All runs reuse the same frozen HTG memory while replacing the LLM action backend for local-node selection and CUDA implementation generation.
As shown in Table~\ref{tab:llm_backend}, HTAM remains effective across APIs, but performance still depends on backend code-generation capability: DS-V4-Flash is weaker, while DeepSeek-R1 and gpt-4o achieve comparable GeoM speedups of $2.003\times$ and $2.037\times$.
This suggests that HTAM provides reusable optimization guidance across APIs, but still requires a capable LLM to produce executable inline CUDA code. Detailed task IDs and API settings are provided in Appendix~\ref{app:llm_backend_details}.

\paragraph{Cross Benchmark Transfer Generalization.}
\label{sec:robustkbench_transfer}

To examine whether HTAM's memory captures transferable optimization experience beyond KernelBench, we evaluate it on five representative Robust-KBench~\citep{lange2025towards} tasks.
Robust-KBench provides workload-oriented kernel tasks with task-specific configurations and forward/backward evaluation, which differ from KernelBench-style operator instances and serve as an independent transfer testbed.
We compare cold-start HTAM, which builds HTG through online exploration on Robust-KBench, with a transfer variant initialized by KernelBench-built HTG memory.
As shown in Table~\ref{tab:robustkbench_transfer}, KernelBench memory improves GeoM speedup from $1.20\times$ to $1.58\times$, with large gains on normalization operators such as LayerNorm and RMSNorm.
In addition, the drop for the Cross-Entropy indicates that transfer remains operator-dependent when the source and target structures mismatch.
Overall, these results suggest that HTAM's hierarchical transition memory can guide inline-CUDA optimization across benchmark suites, while still requiring operator-compatible priors.
Details are provided in Appendix~\ref{app:robustkbench_transfer}.

\begin{table}[t]
\centering
\small
\setlength{\tabcolsep}{4pt}
\renewcommand{\arraystretch}{1.10}
\begin{tabular}{@{}lccc@{}}
\toprule
\textbf{Task} 
& \textbf{Cold Spd.$\uparrow$} 
& \textbf{KBMem Spd.$\uparrow$} 
& \textbf{$\Delta$ Spd.$\uparrow$} \\
\midrule
LayerNorm 
& 2.15$\times$ 
& 3.91$\times$ 
& +1.76$\times$ \\

RMSNorm 
& 2.05$\times$ 
& 3.64$\times$ 
& +1.59$\times$ \\

Cross Entropy 
& 0.62$\times$ 
& 0.52$\times$ 
& -0.10$\times$ \\

Linear 
& 0.92$\times$ 
& 1.09$\times$ 
& +0.17$\times$ \\

Linear + ReLU 
& 1.01$\times$ 
& 1.22$\times$ 
& +0.21$\times$ \\
\midrule
\multicolumn{1}{@{}>{\columncolor{gray!12}[0pt][\tabcolsep]}l}{\textbf{GeoM Spd.}}
& \multicolumn{1}{>{\columncolor{gray!12}[\tabcolsep][\tabcolsep]}c}{1.20$\times$}
& \multicolumn{1}{>{\columncolor{gray!12}[\tabcolsep][\tabcolsep]}c}{\textbf{1.58$\times$}}
& \multicolumn{1}{>{\columncolor{gray!12}[\tabcolsep][0pt]}c@{}}{\textbf{+0.38$\times$}} \\
\bottomrule
\end{tabular}
\caption{
Cross-benchmark memory transfer on five Robust-KBench tasks. Cold denotes no KernelBench memory, and KBMem denotes initialization with KernelBench-built HTG memory. Speedups are in $\times$.
}
\label{tab:robustkbench_transfer}
\end{table}

\subsection{Ablation Study}

We ablate three key design choices of HTAM: hierarchical organization, prefix-aware transition, and multi-step evolution. All variants use the same evaluation setting and differ only in the ablated or replaced component, allowing us to isolate the contribution of each mechanism.

For \textbf{Hierarchical Organization}, we remove the global--local decision structure in~\eqref{eq:HTG}, remove the local memory used for candidate construction in~\eqref{eq:local nodes}, and further test a randomized-memory variant.
For \textbf{Prefix-aware Transition}, we ablate the transition rule in~\eqref{eq:global direction} by selecting global directions without the optimization prefix $p_g$ in~\eqref{eq:prefix_direction} or without the position-aware weight in~\eqref{eq:prefix_weight}.
For \textbf{Evolution Strategy}, we freeze memory updates after initializing the HTG, and vary the number of evolution steps $T$ to test whether the gains rely on multi-step evolution.
All results are summarized in Table~\ref{tab:ablation}, with implementation details in Appendix~\ref{app:ablation_details}.

\begin{table}[t]
\centering
\footnotesize
\setlength{\tabcolsep}{1.5pt}
\renewcommand{\arraystretch}{1.08}
\begin{tabular*}{\columnwidth}{@{\extracolsep{\fill}}lccc}
\toprule
\textbf{Variant} 
& \textbf{Corr.$\uparrow$} 
& \textbf{GeoM$\uparrow$} 
& \textbf{$\Delta$GeoM} \\
\midrule

\rowcolor{gray!12}
\multicolumn{4}{c}{\textbf{Full Method}} \\
\midrule
HTAM ($T=6$)
& \textbf{98.4\%} & \textbf{1.978$\times$} & 0 \\

\midrule
\rowcolor{gray!12}
\multicolumn{4}{c}{\textbf{Hierarchical Organization}} \\
\midrule
w/o hierarchy structure \eqref{eq:HTG}
& 89.6\% & 1.464$\times$ & -0.514$\times$ \\
w/o local memory \eqref{eq:local nodes}
& 71.6\% & 0.974$\times$ & -1.004$\times$ \\
randomized memory 
& 88.8\% & 0.812$\times$ & -1.166$\times$ \\

\midrule
\rowcolor{gray!12}
\multicolumn{4}{c}{\textbf{Prefix-Aware Transition}} \\
\midrule
w/o prefix $p_{g}$ \eqref{eq:prefix_direction} 
& 94.4\% & 1.683$\times$ & -0.295$\times$ \\
w/o position-aware weight~\eqref{eq:prefix_weight}  
& 95.6\% & 1.787$\times$ & -0.191$\times$ \\

\midrule
\rowcolor{gray!12}
\multicolumn{4}{c}{\textbf{Evolution Strategy}} \\
\midrule
w/o updated memory \eqref{eq:memory_update} 
& 96.4\% & 1.842$\times$ & -0.136$\times$ \\
1-step evolution  
& 91.2\% & 1.428$\times$ & -0.550$\times$ \\
3-step evolution 
& 95.2\% & 1.756$\times$ & -0.222$\times$ \\

\bottomrule
\end{tabular*}
\caption{
Ablation study of HTAM on the full 250-task KernelBench suite.
$\Delta$GeoM denotes the change in geometric mean speedup relative to the full HTAM setting.
}
\label{tab:ablation}
\end{table}

As shown in Table~\ref{tab:ablation}, the largest drops come from the \textbf{Hierarchical Organization} group.
Removing the global--local hierarchy reduces GeoM from $1.978\times$ to $1.464\times$, and removing local memory further drops it to $0.974\times$.
The randomized-memory variant reaches only $0.812\times$, showing that HTAM relies on structured global--local experience rather than simply adding textual memory.

The \textbf{Prefix-Aware Transition} ablations also reduce performance.
Removing the optimization prefix $p_g$ lowers GeoM to $1.683\times$, while removing the position-aware weight gives $1.787\times$.
This confirms that global direction selection should depend on recent optimization history.

The \textbf{Evolution Strategy} ablations highlight multi-step memory use.
Freezing memory updates lowers GeoM to $1.842\times$, and reducing evolution depth to 1 or 3 steps gives $1.428\times$ and $1.756\times$, respectively.
Together, these results support HTAM's central design: hierarchical memory organizes experience, prefix-aware transition guides the next direction, and multi-step evolution turns feedback into reusable optimization knowledge.

\subsection{Case Study}
\label{sec:case_study}

We examine a Swish activation operator (\texttt{kb\_l1\_25\_swish}) to illustrate how HTAM turns memory-guided decisions into executable inline CUDA code.
Starting from the PyTorch implementation \texttt{x * torch.sigmoid(x)}, HTAM follows the Data Reuse $\rightarrow$ Memory Access $\rightarrow$ Memory Access global sequence and progressively applies Light fusion, Guarded \texttt{float4} vectorization, and Read-only-load refinement.
As shown in Figure~\ref{fig:swish_case_main}, these steps reduce runtime from 18.2\,ms to 7.76\,ms, improving the speedup from $1.00\times$ to $2.35\times$.

This case shows that HTAM does not merely retrieve high-level optimization hints.
Instead, it uses hierarchical memory to guide concrete inline-CUDA edits, producing auditable kernels with explicit guards, vectorized hot loops, and executable fallback logic.
The complete trajectory and full implementation are provided in Appendix~\ref{app:case_study_swish}.

\begin{figure}[!htbp]
\centering
\small

\setlength{\tabcolsep}{2.4pt}
\renewcommand{\arraystretch}{1.06}
\begin{tabular*}{0.98\columnwidth}{@{\extracolsep{\fill}}c l l r r@{}}
\toprule
\textbf{Step} 
& \textbf{Global direction} 
& \textbf{Local strategy} 
& \textbf{Time}
& \textbf{Spd.} \\
\midrule
Base & \multicolumn{2}{l}{-----PyTorch Reference Code-----} 
& 18.2 ms & 1.00$\times$ \\
1 & Data Reuse & Light fusion 
& 9.56 ms & 1.90$\times$ \\
2 & Memory Access & Guarded f4 path 
& 7.77 ms & 2.34$\times$ \\
3 & Memory Access & Read-only load 
& 7.76 ms & 2.35$\times$ \\
\bottomrule
\end{tabular*}

\vspace{0.55em}

\begin{lstlisting}[style=cudaCase]
__global__ void swish4(const float4* x, float4* y, int n) {
  for (int i = blockIdx.x * blockDim.x + threadIdx.x;
       i < n; i += blockDim.x * gridDim.x) {
    float4 v = __ldg(x + i);
    v.x = v.x / (1.f + __expf(-v.x));
    v.y = v.y / (1.f + __expf(-v.y));
    v.z = v.z / (1.f + __expf(-v.z));
    v.w = v.w / (1.f + __expf(-v.w));
    y[i] = v;
  }
}
\end{lstlisting}

\vspace{-0.45em}

\caption{
Swish case study. The table shows step-wise strategies and speedups; f4 denotes \texttt{float4}. The code highlights the final \texttt{float4} hot loop.
}
\label{fig:swish_case_main}
\end{figure}

\section{Conclusion}

In this work, we introduce HTAM, a hierarchical transition-attended memory framework for LLM-based operator optimization. HTAM organizes optimization experience into global directions, local strategies, and transition memories, enabling a coarse-to-fine decision process from what to optimize to how to implement it. We further design a transition-aware, attention-inspired selection strategy that connects the current decision with recent optimization history, providing reusable multi-step guidance with compact pairwise edge memory.

Our experiments on KernelBench show that HTAM consistently improves correctness, fast-solution rate, and speedup over vanilla LLM generation and iterative refinement baselines. These results suggest that structured memory can serve not only as a repository of past optimization cases, but also as a decision prior for selecting the next useful optimization direction.


\section*{Limitations}

Our evaluation is limited by computational and API costs. 
HTAM is evaluated mainly on KernelBench and additionally on five representative Robust-KBench tasks under a fixed search budget. 
We do not exhaustively sweep all base LLMs, decoding budgets, sampling strategies, memory configurations, or hardware platforms. 
Therefore, the reported results should be interpreted as evidence for HTAM under a controlled setting, while its generalization to broader APIs, GPUs, and deployment environments remains to be further tested.
Extending HTAM to substantially different hardware or operator families may also require adapting the memory schema and validation protocol, although node and edge evidence are updated from executable trajectories.

HTAM also relies on executable validation rather than formal verification. 
The evaluator checks compilation, correctness, and runtime behavior on benchmark-provided inputs, which is appropriate for our experimental setting but does not prove correctness for all possible shapes, data distributions, or integration contexts. 
Future work could extend HTAM to broader hardware platforms, richer operator families, and more rigorous validation settings.

\section*{Acknowledgments}

This work is supported by the Zhongguancun Academy (Grant No.s C20250203) and the National Natural Science Foundation of China (Grant No. 62506365).

\bibliography{custom}


\appendix

\section{Transition Memory Design}
\label{app:graph_memory_design}

\subsection{Design Rationale}
\label{app:transition_design}

A central design choice of HTG is how to store and reuse the transition experience.
In operator optimization, the effect of a decision often depends on previous transformations.
For example, after memory access has been improved, the next useful direction may shift toward instruction throughput or boundary simplification.
Therefore, HTAM should not treat each optimization step as an independent next-action prediction.

A natural solution is to store full optimization chains.
That is, each historical prefix is used as a memory key for selecting the next global direction.
This design is expressive because it can directly model long-range path dependence.
However, it is difficult to use in practice: exact chains rarely repeat across operators, the memory becomes sparse, and retrieving long histories consumes additional prompt budget.

Another simple solution is to store only the transition from the immediately previous global node:
\begin{equation}
Q(g_{t-1}, g).
\end{equation}
This next-step transition memory is compact, but it only uses the last selected direction.
It may miss useful dependencies from earlier steps, such as whether memory access has already been optimized, whether register pressure has increased, or whether boundary handling has introduced correctness risks.

HTAM therefore adopts a middle ground.
It stores transition evidence on pairwise global edges, but aggregates evidence from multiple recent global nodes when selecting the next direction:
\begin{equation}
\operatorname{score}_g(g \mid p_t^g, s_t)
=
\sum_{i<t}\alpha_i\,\phi(g_i,g\mid s_t),
\end{equation}
where $p_t^g=(g_1,\ldots,g_{t-1})$ is the recent global-node prefix, $\phi(g_i,g\mid s_t)$ is the state-conditioned transition utility stored on the edge from previous node $g_i$ to candidate node $g$, and $\alpha_i$ assigns larger weights to more recent decisions.

This design separates storage from decision-time aggregation.
HTAM does not store a separate memory for every full chain.
Instead, it stores reusable pairwise edge memories and combines them according to the recent prefix at decision time.
Thus, global edges serve as transition-level memory: they record which optimization directions tend to follow, complement, or conflict with each other, while avoiding the cost of storing entire trajectories.

\subsection{Cost Analysis}
\label{app:transition_cost}

The main cost difference comes from the granularity of the memory key.
With full-chain memory, every prefix--direction pair $(p_t^g,g)$ becomes a separate key.
Given $|V_G|$ global nodes and search horizon $T$, the number of possible transition keys grows as
\begin{equation}
\sum_{t=1}^{T}|V_G|^t = O(|V_G|^T).
\end{equation}
This growth is problematic for three reasons.
First, storage cost increases rapidly as the horizon grows.
Second, evidence becomes sparse because most long prefixes appear only a few times.
Third, reuse becomes difficult because two operators may share useful partial decisions but rarely share the exact same full chain.

In contrast, HTAM stores transition memory only on directed global edges.
The number of edge memories is
\begin{equation}
|V_G|^2,
\end{equation}
which is independent of the search horizon.
At decision time, HTAM recovers prefix awareness by aggregating multiple edge memories from the recent global-node prefix.
Therefore, the memory remains compact, while the decision can still depend on more than the immediately previous node.

Table~\ref{tab:transition-memory-design} summarizes the comparison.
Full-chain memory is expressive but difficult to populate and reuse.
Next-step memory is compact but loses longer-range context.
HTAM keeps the same storage order as pairwise transition memory, while using prefix-aware aggregation to model multi-step dependencies.

\begin{table}[t]
\centering
\footnotesize
\setlength{\tabcolsep}{3pt}
\renewcommand{\arraystretch}{1.12}
\begin{tabularx}{\columnwidth}{
    >{\raggedright\arraybackslash}p{0.31\columnwidth}
    >{\raggedright\arraybackslash}p{0.24\columnwidth}
    >{\raggedright\arraybackslash}X
}
\toprule
\textbf{Memory design} & \textbf{Storage cost} & \textbf{Property} \\
\midrule
Full-chain memory
& $O(|V_G|^T)$
& Captures full path dependence, but is sparse and difficult to reuse. \\

Next-step memory
& $O(|V_G|^2)$
& Compact, but only conditions on the immediately previous node. \\

HTAM edge memory
& $O(|V_G|^2)$
& Compact, while aggregating evidence from multiple recent nodes. \\
\bottomrule
\end{tabularx}
\caption{Comparison of transition-memory designs.}
\label{tab:transition-memory-design}
\end{table}

\section{Hierarchical Transition Graph (HTG)}

This appendix describes the initial Hierarchical Transition Graph (HTG) memory used by HTAM.
HTAM starts from an initialized HTG memory bank that contains the global optimization directions, local strategy cards, and directed global-to-global transition edges.

The initialized HTG contains both static priors and writable runtime fields.
Static priors provide reusable optimization knowledge, such as the goal of each global direction, the use conditions of local strategies, and the rationale of global transitions.
During optimization, HTAM operates on a run-specific writable HTG instance and updates global nodes, local nodes, and transition edges with runtime evidence from compilation, correctness, speedup, and retrieval outcomes.
The released initial files are kept unchanged only to make experiments reproducible; the algorithm itself treats the HTG instance as readable and updatable memory.

\begin{table}[!htb]
\centering
\small
\begin{tabularx}{\columnwidth}{lY}
\toprule
\textbf{Item} & \textbf{Description} \\
\midrule
Initial memory name
& \texttt{htam\_graph\_initial} \\

Global memory
& Global optimization directions with static priors and runtime statistics. \\

Local memory
& Local strategy cards with implementation guidance and empirical evidence. \\

Transition memory
& Directed global-to-global edges with transition priors and runtime transition evidence. \\

Update policy
& Each run uses a writable HTG instance. Runtime evidence is written back to this instance, while the released initial files remain unchanged for reproducibility. \\
\bottomrule
\end{tabularx}
\caption{Initial HTG memory used by HTAM.}
\label{tab:htg-memory-overview}
\end{table}

\subsection{Global Nodes}
\label{app:global-nodes}

Global nodes store direction-level optimization memory in HTG.
Each global node corresponds to a high-level optimization intent, such as improving memory access, increasing data reuse, simplifying boundary cases, or adjusting thread/block mapping.
These nodes provide the coarse-grained decision space for HTAM: before applying a concrete optimization strategy, HTAM first selects a global node that best matches the current operator state and recent optimization history.

Each global node contains both initial priors and runtime evidence.
The initial priors describe the optimization goal and the types of operator bottlenecks that the node is intended to address.
During optimization, the node is updated with empirical evidence from evaluated kernels, including whether this direction tends to improve compilation success, correctness, runtime speedup, or downstream transition quality.
Thus, global nodes are not fixed labels; they are writable memory units that accumulate direction-level experience across optimization trajectories.

Table~\ref{tab:global-nodes} lists the global optimization intents used to initialize the HTG memory.

\begin{table}[t]
\centering
\footnotesize
\setlength{\tabcolsep}{3pt}
\renewcommand{\arraystretch}{1.12}
\begin{tabularx}{\columnwidth}{
    >{\raggedright\arraybackslash}p{0.31\columnwidth}
    >{\raggedright\arraybackslash}X
}
\toprule
\textbf{Short label} & \textbf{Goal} \\
\midrule
Memory Access
& Improve coalesced access, stride patterns, and effective memory throughput. \\

Boundary
& Isolate boundary cases so that the hot path becomes simpler and more predictable. \\

Throughput
& Improve effective throughput through safe unrolling, blocking, hoisting, or related transformations. \\

Data Reuse
& Improve local reuse through tiling, staging, or lightweight fusion. \\

Parallel Mapping
& Align block/thread mapping with output layout and hardware execution behavior. \\
\bottomrule
\end{tabularx}
\caption{Global optimization intents in the initial HTG memory.}
\label{tab:global-nodes}
\end{table}

\subsection{Global Transition Edge}
\label{app:global-edge-init}

HTAM initializes transition-edge memory over the complete directed global pair
space. Specifically, for every ordered pair of global directions
$(g_{\mathrm{src}}, g_{\mathrm{dst}})\in V_G\times V_G$, HTAM creates a
transition memory object $E_G(g_{\mathrm{src}}, g_{\mathrm{dst}})$. This includes
same-intent transitions, since an optimization trajectory may keep refining
under the same global direction across consecutive steps.

Each transition edge stores two types of information. First, it contains a
static transition prior, including the source and destination intents, the
expected transition rationale, typical trigger conditions, and potential risks.
Second, it maintains runtime feedback statistics accumulated from the executable
trajectories, such as compile outcomes, correctness outcomes, runtime gains, and
state-conditioned transition evidence.

For common and high-confidence transitions, HTG provides more
specific initial descriptions and priors. All remaining directed pairs are still
explicitly initialized, but start from default transition priors and empty
runtime statistics. During optimization, these edges are updated in the same
way once the corresponding transition is observed in a trajectory. Therefore,
the transition lookup $E_G(g_i,g_t)$ used by global routing is always defined
for any previous global direction $g_i$ and candidate direction $g_t$.

At the first evolution step, the global prefix is empty, and no transition edge
is read. Transition-edge feedback is updated from the second step onward, when
both the previous global direction $g_{t-1}$ and the current direction $g_t$
are available. Local-to-local edges are not used by the active scoring logic.

\subsection{Local Nodes}
\label{app:local-nodes}

Local nodes store implementation-level optimization memory under each global node.
While a global node represents a high-level optimization direction, a local node records a more concrete strategy that can be selected after the global direction is chosen.
Each local node contains prior knowledge about the strategy, its applicable conditions, expected benefits, common risks, and runtime evidence accumulated from evaluated kernels.
In this way, HTAM does not retrieve an unconstrained text memory, but narrows the search from a global optimization intent to a small set of structured local memories.

Table~\ref{tab:local-node-memory} summarizes the types of information maintained in local nodes.
The exact implementation may store these fields in structured JSON files or run-local auxiliary files, but logically they belong to the same writable HTG memory instance.

\begin{table}[t]
\centering
\footnotesize
\setlength{\tabcolsep}{3pt}
\renewcommand{\arraystretch}{1.12}
\begin{tabularx}{\columnwidth}{
    >{\raggedright\arraybackslash}p{0.31\columnwidth}
    >{\raggedright\arraybackslash}X
}
\toprule
\textbf{Memory type} & \textbf{Content} \\
\midrule
Strategy prior
& Describes the local optimization strategy and the type of operator state it is intended to improve. \\

Applicability memory
& Records when the strategy is likely to be useful and when it should be avoided. \\

Risk memory
& Stores common failure modes, correctness constraints, and cases where the strategy may hurt performance. \\

Runtime evidence
& Accumulates empirical feedback from evaluated kernels, including compile results, correctness outcomes, speedup changes, and failure observations. \\

Retrieval summary
& Maintains compact summaries of useful past experience so that later steps can reuse successful or failed attempts under similar conditions. \\
\bottomrule
\end{tabularx}
\caption{Information maintained in local nodes of the HTG memory.}
\label{tab:local-node-memory}
\end{table}

The initial HTG memory contains local nodes under each global optimization intent.
For example, memory-access optimization includes local memories for vectorized loading, read-only cache usage, stride-aware access restructuring, and register-level micro-tiling.
Boundary simplification includes local memories for fast-path predicates, tail isolation, and shape-specific dispatch.
Throughput optimization includes local memories for small-factor unrolling, register accumulation, and invariant hoisting.
Data-reuse optimization includes local memories for shared-memory tiling, staged reuse, and lightweight fusion.
Parallel-mapping adjustment includes local memories for output-aligned block shapes, warp-lane remapping, and thread coarsening.

During optimization, HTAM updates these local nodes with new evidence after each evaluated kernel.
Successful local decisions strengthen the corresponding memory with positive runtime evidence, while failed attempts add negative evidence such as invalid preconditions, correctness violations, or performance regressions.
Thus, local nodes serve as the fine-grained memory layer of HTG: they preserve concrete optimization experience without exposing the model to an unstructured and overly large memory bank.

\subsection{Memory Organization in HTG}
\label{app:htg-memory-contents}

Table~\ref{tab:htam_memory_objects} summarizes the representative memory maintained by the three core components of HTG.
The table illustrates the abstraction used in the main text and does not enumerate all implementation fields.

HTG separates each memory object into two complementary parts: prior memory and runtime memory.
Prior memory provides the initial optimization knowledge used before task-specific feedback is available.
It describes what a direction, strategy, or transition is intended to capture, together with typical triggers, expected benefits, and potential risks.
Runtime memory is accumulated during optimization from executable trajectories, including compilation outcomes, correctness results, speedup changes, and failure observations.
This separation allows HTAM to start from reusable optimization knowledge while continuously adapting the memory to the evidence observed during search.

The three memory components play different roles.
Global nodes summarize direction-level experience, local nodes preserve strategy-level evidence, and global transition edges record path-dependent relations between directions.
Together, they allow HTAM to reuse optimization experience at multiple granularities without storing complete trajectories or relying on a flat collection of unstructured textual memories.

\begin{table}[!htb]
\centering
\small
\setlength{\tabcolsep}{3pt}
\renewcommand{\arraystretch}{1.15}
\begin{tabularx}{\columnwidth}{@{}p{0.25\columnwidth}XX@{}}
\toprule
\textbf{Memory object} & \textbf{Prior memory} & \textbf{Runtime memory} \\
\midrule
Global node
& Optimization goal; typical triggers; applicable patterns; risks; expected gain types.
& Attempt frequency; success trend; aggregated gain summary; retrieval helpfulness. \\

Local node
& Use/avoid conditions; code patterns; edit recipe; invariants; verification checklist; common failure modes.
& Positive and negative evidence; empirical strategy feedback; compact case summaries. \\

Global transition edge
& Transition rationale; pivot conditions; expected benefits; risks; avoidance cases.
& Transition gain statistics; compilation and correctness outcomes; context-conditioned feedback. \\
\bottomrule
\end{tabularx}
\caption{Representative memory maintained by HTG components. Global nodes store direction-level memory, local nodes store strategy-level memory, and global transition edges store transition-level memory.}
\label{tab:htam_memory_objects}
\end{table}

\subsection{State Summarization}
\label{app:state_fields}

HTAM models operator optimization as a sequential decision process.
At step $t$, the decision state $s_t$ summarizes the information needed to choose the next optimization action.
It combines environment-derived signals from the current implementation and executable feedback with memory-derived signals retrieved from HTG.
We use $s_t$ as a compact decision summary rather than a full record of all logs, code versions, or implementation fields.

\begin{table}[t]
\centering
\footnotesize
\setlength{\tabcolsep}{3pt}
\renewcommand{\arraystretch}{1.12}
\begin{tabularx}{\columnwidth}{
    >{\raggedright\arraybackslash}p{0.32\columnwidth}
    >{\raggedright\arraybackslash}X
}
\toprule
\textbf{Signal group} & \textbf{Information included in $s_t$} \\
\midrule

Task and code context
& Benchmark task, operator type, input shape, current implementation, and a concise code summary. \\

Executable feedback
& Compile status, correctness status, runtime, speedup, timeout, and dominant failure signal. \\

Search progress
& Best-so-far result, recent improvement, stagnation indicator, and recent global/local optimization path. \\

Code symptoms
& Observable symptoms such as strided access, branch-heavy loops, repeated index computation, memory-bound behavior, or boundary-handling risks. \\

\bottomrule
\end{tabularx}
\caption{Compact summary of the decision state $s_t$ used by HTAM. The state combines environment-derived feedback with memory-derived signals from global nodes, local nodes, and transition edges.}
\label{tab:state_fields}
\end{table}

\subsection{Global Direction Selection Details}
\label{app:transition_reward}

This appendix specifies the lightweight scoring function $f(\cdot)$ used to compute the transition utility $\phi(g_i,g_t\mid s_t)$ in Eq.~\eqref{eq:score}.
Given the current state $s_t$, a previous global direction $g_i$, and a candidate direction $g_t$, HTAM reads the transition-edge memory $E_G(g_i,g_t)$ and extracts a state-matched feature vector.
If a state-conditioned bucket has enough observations, the bucket statistics are used; otherwise, HTAM falls back to aggregate edge statistics.
No additional LLM call is used in this step.

The scoring function is implemented as a fixed linear reward over transition-memory features:
\begin{equation}
\phi(g_i,g_t\mid s_t)
=
\mathbf{w}^{\top}\mathbf{z}(s_t,g_i,g_t),
\label{eq:transition_phi}
\end{equation}
The feature vector $\mathbf{z}$ is decomposed into improvement, reliability, and risk-related components:
\begin{equation}
\small
\begin{aligned}
\mathbf{z}
&=
[\mathbf{z}_{\mathrm{imp}},\mathbf{z}_{\mathrm{rel}},\mathbf{z}_{\mathrm{risk}}], \\
\mathbf{z}_{\mathrm{imp}}
&=
[\bar r_{\mathrm{imm}}, \bar r_{\mathrm{fut}}, p_{\mathrm{pos}}], \\
\mathbf{z}_{\mathrm{rel}}
&=
[p_{\mathrm{succ}}, p_{\mathrm{comp}}, p_{\mathrm{corr}},
p_{\mathrm{safe}}, h_{\mathrm{ctx}}], \\
\mathbf{z}_{\mathrm{risk}}
&=
[-p_{\mathrm{cfail}}, -p_{\mathrm{corfail}},
-p_{\mathrm{neg}}, -\rho_{\mathrm{risk}}].
\end{aligned}
\label{eq:transition_feature_vector}
\end{equation}

Here, $\mathbf{z}_{\mathrm{imp}}$ measures improvement evidence, where
$\bar r_{\mathrm{imm}}$ and $\bar r_{\mathrm{fut}}$ are the average immediate and future log-runtime gains, and $p_{\mathrm{pos}}$ is the positive-outcome rate.
The reliability component $\mathbf{z}_{\mathrm{rel}}$ includes success, compile-pass, correctness-pass, and safe-improvement rates, together with $h_{\mathrm{ctx}}$, which indicates whether state-conditioned statistics are available.
The risk component $\mathbf{z}_{\mathrm{risk}}$ contains negative signals, including compile-failure rate, correctness-failure rate, executable-negative rate, and the risk estimate $\rho_{\mathrm{risk}}$.
Therefore, the reward favors transitions that have produced reliable improvements under similar states and penalizes transitions that repeatedly lead to failures or regressions.

\subsection{Local Strategy Selection Details}
\label{app:local_selection_prompt}

After the global direction $g_t$ is selected, HTAM restricts the local strategy space to the local nodes associated with $g_t$, as defined in Eq.~\eqref{eq:local nodes}.
This step does not search over all local memories.
Instead, it asks the LLM to compare only the candidate local nodes under the selected global direction and choose the one that best matches the current state and implementation.

The local-selection prompt contains four types of information: the current decision state $s_t$, the current implementation $c_t$, the selected global-node memory $V_G(g_t)$, and the candidate local-node memories $\{V_{GL}(l)\}_{l\in\mathcal{L}(g_t)}$.
The state $s_t$ summarizes the current optimization stage, executable feedback, recent trajectory, and failure or stagnation signals.
The selected global node provides direction-level context, such as the optimization goal, typical triggers, and risks.
Each candidate local node provides strategy-level memory, including use conditions, avoid conditions, relevant code symptoms, expected benefits, correctness constraints, and empirical feedback.

The LLM is instructed to select exactly one local node and return a compact structured output:
\begin{table}[H]
\centering
\footnotesize
\setlength{\tabcolsep}{3pt}
\renewcommand{\arraystretch}{1.08}
\begin{tabularx}{\columnwidth}{
    >{\raggedright\arraybackslash}p{0.42\columnwidth}
    >{\raggedright\arraybackslash}X
}
\toprule
\textbf{Field} & \textbf{Meaning} \\
\midrule
\texttt{selected\_local\_node}
& Selected local memory node. \\

\texttt{rationale}
& Short reason for the selection. \\

\texttt{edit\_plan}
& High-level plan for generating the next implementation. \\
\bottomrule
\end{tabularx}
\caption{Output schema of the local-node selection prompt.}
\label{tab:local_selection_schema}
\end{table}

The selected local node and edit plan are then passed to the code-generation stage.
If the LLM returns an invalid local-node identifier, HTAM falls back to the highest-ranked valid local node under the same global direction.
If no valid local node exists under the selected global node, the step is treated as invalid.

This design separates transition-aware global routing from local edit planning.
The global direction is selected by deterministic HTG scoring, while the LLM is only used to interpret the current code and choose among a small set of relevant local memories.

\section{Detailed Experimental Protocol}
\label{app:exp_protocol}

This appendix provides protocol details for the experimental setup and main comparison in Section~\ref{sec:experiments}.
The main text summarizes the benchmark, hardware, reference runtime, and baseline groups; here, we provide the task coverage, environment, metric definitions, search budgets, baseline definitions, and comparability notes for reported external systems.

\subsection{Hardware and Software Environment}
\label{app:env_protocol}

All reproduced experiments are conducted on a single server with 8 NVIDIA A100-SXM4-80GB GPUs, an Intel Xeon Platinum 8362 CPU, and 935\,GiB RAM.
The software environment uses Ubuntu 22.04.5 LTS, CUDA 12.8, PyTorch 2.9.1+cu128, and KernelBench package 0.2.0.dev0 at commit \texttt{423217d}.
We evaluate the full 250-task KernelBench suite, including 100 Level-1 tasks, 100 Level-2 tasks, and 50 Level-3 tasks.

Unless otherwise specified, HTAM and all reproduced search, refinement, and memory baselines use DeepSeek-R1 as the base LLM, seed 42, and the same task set, prompts, budgets, method-specific initialized memory assets, and executable validation protocol.
For API-based LLM calls, exact token-level reproducibility may still be affected by provider-side nondeterminism; therefore, we release task IDs, prompts, configurations, and run manifests to support reproducibility.
For Direct Generation rows in Table~\ref{tab:main_kernelbench_leaderboard}, we additionally evaluate DeepSeek-V3, DeepSeek-V4-Flash, and Gemini-2.5-Flash-Thinking.

\begin{table}[!htb]
\centering
\small
\setlength{\tabcolsep}{4pt}
\renewcommand{\arraystretch}{1.08}
\begin{tabularx}{\columnwidth}{lX}
\toprule
\textbf{Field} & \textbf{Value} \\
\midrule
GPU & 8$\times$ NVIDIA A100-SXM4-80GB \\
CPU & Intel Xeon Platinum 8362 \\
RAM & 935\,GiB \\
OS & Ubuntu 22.04.5 LTS \\
CUDA & 12.8 \\
PyTorch & 2.9.1+cu128 \\
KernelBench & package 0.2.0.dev0, commit \texttt{423217d} \\
Task coverage & 250 tasks: L1=100, L2=100, L3=50 \\
Default reproduced model & DeepSeek-R1 \\
Additional direct-generation models & DeepSeek-V3, DeepSeek-V4-Flash, Gemini-2.5-Flash-Thinking \\
Temperature & 0.2 \\
Code-generation max tokens & 16384 \\
Local-selection max tokens & 2048, used by HTAM local-node selection \\
Random seed & 42 \\
\bottomrule
\end{tabularx}
\caption{Hardware, software, benchmark, and inference settings for reproduced experiments.}
\label{tab:env_details}
\end{table}

For timing stability, tasks are distributed across available GPUs while avoiding multiple active evaluator jobs on the same GPU whenever possible.
All reproduced rows use the same executable evaluation pipeline and hardware settings unless explicitly noted otherwise.
Reported external systems are not covered by this environment table because they may use different hardware, base models, budgets, and evaluation protocols.

\subsection{Evaluation Metrics and Semantics}
\label{app:metric_semantics}

For KernelBench, speedup is defined as
\[
    \mathrm{speedup}
    =
    \frac{T_{\mathrm{torch\ reference}}}{T_{\mathrm{candidate}}},
\]
where $T_{\mathrm{torch\ reference}}$ is the PyTorch eager reference runtime and $T_{\mathrm{candidate}}$ is the runtime of the generated candidate.
For Robust-KBench, we follow its native convention and report speedup against the torch-native runtime.

A candidate is counted as compiled if it passes compilation, and as correct only if it also passes correctness checking.
Fast@1 denotes the percentage of tasks whose best candidate is both correct and faster than the corresponding PyTorch reference.
When threshold metrics are reported, Valid@$\rho$ denotes the percentage of tasks whose best candidate is correct and reaches at least $\rho\times$ speedup.
GeoMean, median, mean, maximum, and percentile speedups are computed from task-level best candidates under the corresponding benchmark denominator.

\begin{table}[!htb]
\centering
\small
\setlength{\tabcolsep}{4pt}
\begin{tabularx}{\columnwidth}{lX}
\toprule
Item & Definition \\
\midrule
KernelBench speedup & $T_{\mathrm{torch\ reference}}/T_{\mathrm{candidate}}$ \\
KernelBench reference & PyTorch eager \\
Robust-KBench speedup & Runtime relative to torch-native reference \\
Compile & Candidate compiles successfully \\
Correct & Compile success and correctness success \\
Fast@1 & Correct and speedup $>1.0$ \\
Valid@$\rho$ & Correct and speedup $\ge \rho$ \\
Speedup summaries & Aggregated from task-level speedups \\
Main semantics & Using the best candidate within budget \\
\bottomrule
\end{tabularx}
\caption{Metric definitions and denominator conventions.}
\label{tab:metric_semantics}
\end{table}

\subsection{External Reported Results}
\label{app:reported_systems}

Table~\ref{tab:main_kernelbench_leaderboard} includes OpenAI-o3, KernelBlaster, and CudaForge as external reported results.
These rows are not strictly controlled, reproduced baselines, since they may use different base models, hardware, search budgets, and evaluation harnesses.
We therefore preserve their reported metric semantics and mark them with superscripts in the main table.

The OpenAI-o3 row is taken from the CudaForge report and is included to contextualize direct generation with a stronger proprietary model.
KernelBlaster and CudaForge are included as reported system-level references.
When an aggregate value is shown, it is computed from the reported level-wise statistics using KernelBench level weights.
Thus, these rows should be read as contextual references rather than direct apples-to-apples comparisons with reproduced HTAM and baseline rows.

In Table~\ref{tab:main_kernelbench_leaderboard}, $\dagger$ marks a reported direct-generation result taken from CudaForge, and $\ddagger$ marks reported external system results.

\subsection{Baseline Details}
\label{app:baseline_details}

We compare HTAM with reproduced baselines and reported external systems.
Unless otherwise specified, HTAM and the reproduced search, refinement, and memory baselines use DeepSeek-R1 under the same task set, search budget, executable validation protocol, timing pipeline, and strict FP32 reference setting.
The Direct Generation group additionally includes several base LLMs to separate method effects from base-model effects.
Reported systems are included only as external references because they may use different models, hardware, budgets, validation procedures, and speedup definitions.

\paragraph{Direct Generation.}
Direct Generation is a single-pass LLM kernel generation baseline without search, iterative feedback, or external memory.
The DeepSeek-R1 Direct Generation row serves as the controlled base-model comparison for HTAM.
We also report Direct Generation with additional base LLMs to show that direct prompting alone does not reliably translate model strength into valid acceleration.

\paragraph{Best-of-$N$ Sampling.}
Best-of-$N$ draws $N=5$ independent Direct Generation samples and selects the best valid candidate according to the benchmark reward.
This baseline represents inference-time sampling without iterative feedback or memory.

\paragraph{Feedback-only Refinement.}
Feedback-only Refinement iteratively revises the current candidate using compiler, correctness, and runtime feedback.
It does not maintain structured memory, hierarchical action selection, or transition-aware decision priors.

\paragraph{Flat Memory Retrieval.}
Flat Memory Retrieval augments generation with retrieved historical experience, but removes HTAM's hierarchical global--local strategy structure and transition-aware edge memory.
It is used to test whether flat retrieval alone is sufficient for multi-step operator optimization.

\paragraph{HTAM.}
HTAM uses hierarchical global--local strategy selection, transition-aware memory, retrieval-as-action, and dynamic memory updates.
It is evaluated under the same executable protocol as the reproduced baselines.

\subsection{Main-Table Notation and Comparability Notes}
\label{app:main_table_notes}

The main comparison table uses compact notation to keep the table caption short. 
Direct Generation denotes single-pass LLM kernel generation without search, iterative execution feedback, or external memory. 
Reproduced search, refinement, memory, and HTAM rows are evaluated under the controlled protocol described above unless otherwise specified.

Rows marked with $^\dagger$ are reported from prior work and may use different base models, hardware, budgets, validation pipelines, and speedup definitions. 
Rows marked with $^\ddagger$ use reported level-wise geometric mean speedups. 
The symbol ``--'' indicates that the corresponding metric is not reported by the original paper. 
Therefore, reported rows are included as contextual references, while the reproduced rows provide the controlled comparison under our protocol.

\subsection{Details of LLM Backend Generalization}
\label{app:llm_backend_details}

We evaluate backend sensitivity on a fixed 50-task KernelBench subset with 20 L1, 20 L2, and 10 L3 tasks.
All runs use the same frozen HTG memory constructed with DeepSeek-R1, the same deterministic global-routing procedure, optimizer, search budget, evaluator, prompt structure, and speedup definition.
The replaced backend is the LLM action backend, covering both \texttt{local\_selection} and \texttt{code\_generation}.
That is, each tested backend is responsible for selecting the local node under the chosen global direction and generating the corresponding CUDA implementation, while global direction selection and memory update remain API-free and unchanged.
The subset is selected before backend comparison and uses only task metadata, reference executability, and PyTorch reference runtime; no HTAM, baseline, or LLM-generated result is used for selection.

\begin{table}[!htb]
\centering
\scriptsize
\setlength{\tabcolsep}{3pt}
\renewcommand{\arraystretch}{1.08}
\begin{tabularx}{\columnwidth}{@{}p{0.08\columnwidth}>{\raggedright\arraybackslash}X@{}}
\toprule
\textbf{Level} & \textbf{Selected task suffixes} \\
\midrule

\textbf{L1} &
\path{5_matrix_scalar_multiplication};
\path{10_3d_tensor_matrix_multiplication};
\path{11_4d_tensor_matrix_multiplication};
\path{13_matmul_for_symmetric_matrices};
\path{16_matmul_with_transposed_a};
\path{17_matmul_with_transposed_b};
\path{19_relu};
\path{23_softmax};
\path{25_swish};
\path{37_frobeniusnorm};
\path{40_layernorm};
\path{41_max_pooling_1d};
\path{42_max_pooling_2d};
\path{43_max_pooling_3d};
\path{44_average_pooling_1d};
\path{45_average_pooling_2d};
\path{47_sum_reduction_over_a_dimension};
\path{57_conv_transposed_2d_square_input_square_kernel};
\path{61_conv_transposed_3d_square_input_square_kernel};
\path{71_conv_transposed_2d_asymmetric_input_square_kernel}. \\

\midrule

\textbf{L2} &
\path{1_conv2d_relu_biasadd};
\path{11_convtranspose2d_batchnorm_tanh_maxpool_groupnorm};
\path{12_gemm_multiply_leakyrelu};
\path{15_convtranspose3d_batchnorm_subtract};
\path{17_conv2d_instancenorm_divide};
\path{18_matmul_sum_max_avgpool_logsumexp_logsumexp};
\path{20_convtranspose3d_sum_residualadd_multiply_residualadd};
\path{21_conv2d_add_scale_sigmoid_groupnorm};
\path{22_matmul_scale_residualadd_clamp_logsumexp_mish};
\path{23_conv3d_groupnorm_mean};
\path{24_conv3d_min_softmax};
\path{25_conv2d_min_tanh_tanh};
\path{28_bmm_instancenorm_sum_residualadd_multiply};
\path{32_conv2d_scaling_min};
\path{36_convtranspose2d_min_sum_gelu_add};
\path{40_matmul_scaling_residualadd};
\path{43_conv3d_max_logsumexp_relu};
\path{44_convtranspose2d_multiply_globalavgpool_globalavgpool_mean};
\path{53_gemm_scaling_hardtanh_gelu};
\path{100_convtranspose3d_clamp_min_divide}. \\

\midrule

\textbf{L3} &
\path{1_mlp};
\path{10_resnet101};
\path{11_vgg16};
\path{13_densenet121transitionlayer};
\path{16_densenet201};
\path{19_mobilenetv1};
\path{25_shufflenetunit};
\path{28_visiontransformer};
\path{34_vanillarnnhidden};
\path{39_gru}. \\

\bottomrule
\end{tabularx}
\caption{
Selected KernelBench tasks for the 50-task backend-sensitivity subset.
For compactness, each item omits its level prefix; e.g., an L1 suffix denotes \texttt{kb\_l1\_\{suffix\}}.
}
\label{tab:llm_backend_tasks}
\end{table}

\subsection{Details of Cross-Benchmark Transfer Generalization}
\label{app:robustkbench_transfer}

We evaluate cross-benchmark memory transfer on five representative Robust-KBench tasks: LayerNorm, RMSNorm, Cross Entropy, Linear, and Linear+ReLU.
The goal of this experiment is to test whether the optimization experience accumulated on KernelBench can provide useful priors for a different benchmark suite, rather than only for additional KernelBench-style task instances.

We compare two HTAM settings under the same Robust-KBench evaluation protocol.
In the cold-start setting, HTAM starts from the canonical HTG schema without loading the KernelBench-built runtime memory.
It then performs the standard six-step online optimization process on each Robust-KBench task, during which memory is updated only with feedback observed within the same task budget.
In the transfer setting, HTAM is initialized with the HTG memory constructed from KernelBench before optimizing Robust-KBench tasks.
This setting therefore reuses KernelBench-accumulated global, local, and transition memories as priors, while still following the same six-step online optimization procedure during Robust-KBench evaluation.

No additional Robust-KBench-specific pre-warmup or separate memory-training phase is performed before evaluation.
Both settings use the same optimizer, search budget, number of evolution steps, evaluator, base model, and speedup definition.
Thus, the comparison isolates the effect of initializing HTAM with KernelBench-built memory.

For each task, speedup is measured against the official PyTorch/native reference used by Robust-KBench.
We report per-task speedup and the geometric mean speedup over the five tasks.
The column $\Delta$ Spd. is computed as the absolute difference between the transfer and cold-start speedups:
\begin{equation}
\Delta \mathrm{Spd.} = \mathrm{Spd.}_{\mathrm{KBMem}} - \mathrm{Spd.}_{\mathrm{Cold}}.
\end{equation}
This metric directly measures how much speedup is gained or lost by initializing HTAM with the KernelBench-built memory under the same evaluation protocol.

\subsection{Ablation Details}
\label{app:ablation_details}

The ablation study isolates three key design choices of HTAM: hierarchical organization, prefix-aware transition, and multi-step evolution.
All variants are evaluated on the full 250-task KernelBench suite, using the same task set as the main KernelBench evaluation.
The suite contains 100 Level-1 tasks, 100 Level-2 tasks, and 50 Level-3 tasks.
All variants use the same base LLM, evaluator, timing protocol, strict FP32 reference setting, and speedup definition as the main experiment.
Unless otherwise specified, the maximum number of evolution steps is set to $T=6$.

To make the comparison controlled, all variants start from the same starting HTG memory.
For the full HTAM setting, memory updates are enabled within each optimization trajectory, so feedback from earlier steps can affect later decisions on the same task.
The \textit{w/o updated memory} variant disables these within-trajectory updates.
Across tasks and variants, we do not allow one ablation run to write back into the persistent memory asset used by another run.

\paragraph{Full HTAM.}
The full method keeps the complete HTAM design, including the hierarchical global--local memory structure in~\eqref{eq:HTG}, local-node memory in~\eqref{eq:local nodes}, prefix-aware transition scoring in~\eqref{eq:global direction}, position-aware weighting in~\eqref{eq:prefix_weight}, and online memory update in Algorithm~\ref{alg:htam}.
It uses $T=6$ evolution steps and serves as the reference point for computing $\Delta$GeoM in Table~\ref{tab:ablation}.

\paragraph{Hierarchical Organization.}
This group tests whether HTAM benefits from organizing optimization experience into a global--local structure, rather than simply injecting textual memory into the prompt.

\paragraph{w/o hierarchy structure.}
This variant removes the hierarchical global--local decision structure in~\eqref{eq:HTG}.
Instead of first selecting a global optimization direction and then selecting local strategies under that direction, the model operates over a flattened memory space.
This tests whether the coarse-to-fine decomposition is useful for narrowing the optimization search space.

\paragraph{w/o local memory.}
This variant keeps global direction selection but removes local-node memory from the generation context.
The model can still select a high-level optimization direction, but it no longer receives strategy-level memory under that direction.
This tests whether high-level directions alone are sufficient, or whether HTAM needs local-node memory to translate a global intent into concrete CUDA edits.

\paragraph{randomized memory.}
This variant preserves the memory injection format and the number of injected memory items, but replaces the retrieved memory content with randomly sampled eligible memory entries.
This controls for prompt length and memory formatting, testing whether the gains come from relevant optimization experience rather than simply adding more text to the prompt.

\paragraph{Prefix-Aware Transition.}
This group tests whether the next global direction should depend on recent optimization history and position-aware transition aggregation.

\paragraph{w/o prefix $p_g$.}
This variant removes the optimization prefix $p_g$ in~\eqref{eq:prefix_direction} when selecting the next global direction.
The global decision is therefore made without explicitly conditioning on the recent sequence of selected global directions.
This tests whether path-dependent global-direction history helps HTAM choose better subsequent optimization directions.

\paragraph{w/o position-aware weight.}
This variant keeps the prefix information but removes the position-aware weight in~\eqref{eq:prefix_weight}.
Instead of emphasizing more recent global directions, it aggregates prefix transition evidence without recency-sensitive weighting.
This tests whether the order and recency of previous optimization directions provide useful guidance for global selection.

\paragraph{Evolution Strategy.}
This group tests whether HTAM benefits from active multi-step evolution and memory updates during optimization.

\paragraph{w/o updated memory.}
This variant uses the same initialized HTG memory as HTAM but disables memory updates during each optimization trajectory.
HTAM can still retrieve global and local memory, but evaluation feedback from the current task is not written back to nodes or edges after each step.
This tests whether dynamically updating memory with recent compilation, correctness, and runtime feedback helps guide later optimization steps.

\paragraph{1-step evolution and 3-step evolution.}
These search-depth ablations keep the full HTAM mechanism but reduce the maximum number of optimization steps to 1 or 3.
They test whether HTAM's gains come from a single strong generation step or from accumulating, retrieving, and updating structured memory across multiple optimization steps.

\begin{table}[!htb]
\centering
\footnotesize
\setlength{\tabcolsep}{3pt}
\renewcommand{\arraystretch}{1.10}
\begin{tabularx}{\columnwidth}{
    >{\raggedright\arraybackslash}p{0.34\columnwidth}
    >{\raggedright\arraybackslash}X
}
\toprule
\textbf{Variant} & \textbf{Definition} \\
\midrule

\rowcolor{gray!12}
\multicolumn{2}{c}{\textbf{Full Method}} \\
\midrule
HTAM ($T=6$)
& Full HTAM with hierarchical global--local memory, prefix-aware transition, online memory update, and six-step evolution. \\

\midrule
\rowcolor{gray!12}
\multicolumn{2}{c}{\textbf{Hierarchical Organization}} \\
\midrule
w/o hierarchy structure
& Removes the global--local hierarchy and uses a flattened memory organization instead of the two-level HTG structure. \\

w/o local memory
& Keeps global direction selection but removes local-node memory from the generation context. \\

randomized memory
& Preserves the memory injection format and count, but replaces retrieved memory with randomly sampled eligible entries. \\

\midrule
\rowcolor{gray!12}
\multicolumn{2}{c}{\textbf{Prefix-Aware Transition}} \\
\midrule
w/o prefix $p_t^g$
& Selects global directions without conditioning on the recent global-direction prefix. \\

w/o position-aware weight
& Keeps prefix transition evidence but removes the position-aware weighting over recent directions. \\

\midrule
\rowcolor{gray!12}
\multicolumn{2}{c}{\textbf{Evolution Strategy}} \\
\midrule
w/o updated memory
& Freezes memory within each trajectory and disables runtime feedback updates to global nodes, local nodes, and transition edges. \\

1-step evolution
& Uses the full HTAM mechanism but limits the search to one optimization step. \\

3-step evolution
& Uses the full HTAM mechanism but limits the search to at most three optimization steps. \\

\bottomrule
\end{tabularx}
\caption{
Definitions of ablation variants.
All variants are evaluated on the full 250-task KernelBench suite using the same task set, base LLM, evaluator, timing protocol, and strict FP32 reference setting as the main KernelBench experiment.
}
\label{tab:ablation_variant_details}
\end{table}

\subsection{LLM API Cost Analysis}
\label{app:api_cost}

We further estimate the LLM API cost of HTAM to clarify the practical overhead of the proposed memory-guided optimization process.
This analysis focuses on the \emph{online optimization cost} of HTAM under the default $T=6$ evolution-step setting.
We separate this online cost from one-time graph construction and memory warmup, because the latter is paid only once and can be reused across tasks, benchmark suites, and future runs.

\paragraph{Cost scope.}
At each evolution step, HTAM first selects a global direction through deterministic HTG scoring over node and transition statistics.
This global selection does not call the LLM.
The LLM is then used in two phases: \texttt{local\_selection}, which selects a local strategy card and produces an edit plan, and \texttt{code\_generation}, which generates the executable \texttt{ModelNew.py} implementation with inline CUDA.
If the generated code fails compilation or violates the execution contract, an additional repair call may be triggered.
Finally, memory update is performed by rule-based aggregation of the evaluation feedback and therefore does not introduce additional LLM API calls.
Thus, a typical successful trajectory with $T=6$ contains $6 \times 2 = 12$ LLM calls per task, excluding occasional repair calls.
Global direction selection, transition scoring, and memory update remain deterministic and API-free.
Therefore, the online API cost is reported under the same 12-call scope for all backends unless explicitly stated otherwise.
Table~\ref{tab:api_pipeline} summarizes this pipeline.

\begin{table}[!htb]
\centering
\scriptsize
\setlength{\tabcolsep}{3pt}
\renewcommand{\arraystretch}{1.05}
\begin{tabular}{@{}p{0.30\columnwidth}cp{0.48\columnwidth}@{}}
\toprule
\textbf{Component} & \textbf{API?} & \textbf{Purpose} \\
\midrule
Global sel. & No & HTG scoring/search \\
\texttt{local\_sel.} & Yes & Local card + edit plan \\
\texttt{code\_gen.} & Yes & Full inline-CUDA code \\
Repair & Opt. & Compile/contract fix \\
Memory update & No & Rule-based update \\
\bottomrule
\end{tabular}
\caption{LLM API usage in one HTAM step. A standard $T=6$ trajectory uses 12 LLM calls per task, excluding optional repair.}
\label{tab:api_pipeline}
\end{table}

\paragraph{Pricing and measurement.}
We compute monetary cost from input and output tokens separately.
For the main estimate, we use the DeepSeek-R1 transit price used in our experiments:
\$0.58 per 1M input tokens and \$2.32 per 1M output tokens.
For reference, we also report the cost under GPT-4o pricing
(\$2.25 per 1M input tokens and \$9.00 per 1M output tokens), and a DeepSeek-V4-Flash baseline
(\$0.29 per 1M input tokens and \$0.43 per 1M output tokens).

To make the cost accounting transparent, we profile a completed six-step HTAM trajectory from KernelBench
(\texttt{kb\_l3\_29\_swinmlp}) as a representative breakdown of the online API pipeline.
This trace illustrates the main cost sources in HTAM, including \texttt{local\_selection} and \texttt{code\_generation} calls across all six evolution steps.
For DeepSeek-R1, several \texttt{local\_selection} calls exceeded the timeout; instead of treating them as free, we impute their output tokens using the measured R1 local-selection output from the successful step.
This gives a conservative online cost estimate for the full six-step process.

\begin{table}[!htb]
\centering
\small
\setlength{\tabcolsep}{4pt}
\renewcommand{\arraystretch}{1.08}
\begin{tabular}{lrrr}
\toprule
\textbf{Stage} & \textbf{Input} & \textbf{Output} & \textbf{Total} \\
 & \textbf{/call} & \textbf{/call} & \textbf{/call} \\
\midrule
R1 local selection & 5,665 & 1,618 & 7,283 \\
R1 code generation & 11,495 & 9,496 & 20,991 \\
GPT-4o local selection & 5,588 & 132 & 5,720 \\
GPT-4o code generation & 10,701 & 4,368 & 15,068 \\
V4-Flash local selection & 5,546 & 193 & 5,739 \\
V4-Flash code generation & 11,495 & 4,639 & 16,134 \\
\bottomrule
\end{tabular}
\caption{Average token usage per LLM call in the six-step cost probe. DeepSeek-R1 produces longer code-generation outputs than GPT-4o and V4-Flash, mainly because reasoning tokens are included in the output accounting.}
\label{tab:api_stage_tokens}
\end{table}

\paragraph{Main online-cost estimate.}
Table~\ref{tab:api_cost_task} reports the estimated online API cost per full six-step task and the projection to the full 250-task KernelBench suite.
All rows include the same 12-call HTAM pipeline, consisting of six \texttt{local\_selection} calls and six \texttt{code\_generation} calls, excluding optional repair.
Under DeepSeek-R1 pricing, one complete HTAM trajectory uses about 103k input tokens and 67k output tokens, or 170k total tokens.
This corresponds to about \$0.21 per task and approximately \$54 for 250 tasks.
Under V4-Flash pricing, the corrected 12-call estimate uses about 131k total tokens per task and costs about \$0.042 per task, or approximately \$10.5 for 250 tasks.
Although R1 uses more output tokens than GPT-4o and V4-Flash, its lower unit price keeps the total projected cost substantially below GPT-4o under the same six-step setting.

\begin{table}[!htb]
\centering
\small
\setlength{\tabcolsep}{3.5pt}
\renewcommand{\arraystretch}{1.08}
\begin{tabular}{lrrrr}
\toprule
\textbf{Model} & \textbf{Input} & \textbf{Output} & \textbf{Cost} & \textbf{250 tasks} \\
 & \textbf{/task} & \textbf{/task} & \textbf{/task} & \textbf{cost} \\
\midrule
DeepSeek-R1 & 102,960 & 66,682 & \$0.214 & \$53.6 \\
GPT-4o & 97,731 & 27,000 & \$0.463 & \$115.7 \\
V4-Flash & 102,244 & 28,992 & \$0.042 & \$10.5 \\
\bottomrule
\end{tabular}
\caption{Estimated online LLM API cost for a full six-step HTAM trajectory. All rows include the standard 12-call pipeline with six local-node selection calls and six code-generation calls, excluding optional repair. V4-Flash local-selection tokens are estimated from saved prompts and responses.}
\label{tab:api_cost_task}
\end{table}

\paragraph{Where the cost comes from.}
Across backends, the dominant cost comes from code generation rather than local-node selection.
For DeepSeek-R1, \texttt{local\_selection} costs about \$0.042 per task, while \texttt{code\_generation} costs about \$0.172 per task.
Thus, code generation accounts for roughly 80\% of the online API cost.
This is expected because each code-generation call produces a complete executable implementation, including inline CUDA code, Python wrapping logic, and contract-related checks.
In contrast, global direction selection, transition scoring, and online memory update are performed by graph search and rule-based aggregation, so they do not add LLM calls.

\begin{table}[!htb]
\centering
\small
\setlength{\tabcolsep}{4pt}
\renewcommand{\arraystretch}{1.08}
\begin{tabular}{lrrr}
\toprule
\textbf{R1 stage} & \textbf{Calls/task} & \textbf{Cost/task} & \textbf{Share} \\
\midrule
\texttt{local\_selection} & 6 & \$0.042 & 19.7\% \\
\texttt{code\_generation} & 6 & \$0.172 & 80.3\% \\
\midrule
Total & 12 & \$0.214 & 100.0\% \\
\bottomrule
\end{tabular}
\caption{Stage-level DeepSeek-R1 cost breakdown. Most online API cost comes from generating full inline-CUDA implementations, while the HTG search and memory update themselves are API-free.}
\label{tab:api_cost_breakdown}
\end{table}

\paragraph{One-time graph construction.}
The above numbers estimate the online cost of running HTAM after the hierarchical memory structure is available.
They do not include one-time graph construction, template preparation, or cold-start memory warmup.
These setup costs are conceptually different from online optimization costs: they are not paid once per task, and they can be amortized over the full benchmark and reused in later experiments.
If an implementation uses LLM calls to construct or warm up the graph, the total campaign cost can be written as
\begin{equation}
C_{\mathrm{total}}
=
C_{\mathrm{build}}
+
N \cdot T \cdot
\left(C_{\mathrm{local}} + C_{\mathrm{code}}\right)
+
C_{\mathrm{repair}},
\end{equation}
where $C_{\mathrm{build}}$ is the one-time graph-construction or warmup cost, $N$ is the number of tasks, $T$ is the number of evolution steps, $C_{\mathrm{local}}$ and $C_{\mathrm{code}}$ are the average costs of local selection and code generation per step, and $C_{\mathrm{repair}}$ accounts for occasional repair or retry calls.
Our reported estimate corresponds to the online term
$N \cdot T \cdot (C_{\mathrm{local}} + C_{\mathrm{code}})$ with $T=6$.
This separation avoids underestimating the cost of LLM-assisted cold start while also avoiding charging the reusable graph-construction cost to every task.

\paragraph{Discussion.}
Overall, the online API cost of HTAM is modest but not negligible.
For the full KernelBench suite, the six-step DeepSeek-R1 estimate is approximately \$54 for 250 tasks.
The cost scales approximately linearly with the number of evolution steps and increases when repair or fallback calls are triggered.
However, the memory-specific operations in HTAM---global direction search, transition scoring, and memory update---do not themselves require LLM API calls.
This indicates that the main monetary overhead comes from generating executable CUDA implementations, while the hierarchical memory structure provides reusable guidance without adding substantial online API cost.

\section{Responsible Research Details}
\label{app:responsible_details}

\paragraph{Artifact use, licenses, and intended use.}
This work uses existing research artifacts, including KernelBench, Robust-KBench, PyTorch, CUDA, and LLM backends, only for research evaluation of GPU operator optimization.
We cite the original creators of the benchmarks, models, and baseline systems in the main paper.
We follow the licenses, access conditions, and intended-use requirements specified by the original artifact providers.
The artifacts produced by our work, including code, prompts, configurations, and evaluation scripts, are intended for research on LLM-based operator optimization and reproducibility analysis, rather than direct safety-critical deployment.

\paragraph{Personally identifying information and offensive content.}
The benchmarks used in this work consist of programmatic operator definitions, reference implementations, generated CUDA code, and execution results.
They are not human-subject datasets and do not contain natural-language documents collected from individuals.
We do not collect new personal data, user data, or annotator data.
Accordingly, the data used in our experiments is not expected to contain personally identifying information or offensive human-written content.

\paragraph{Computational budget and model information.}
Our main experiments are conducted under the hardware and software settings described in Section~\ref{sec:experiments} and Appendix~\ref{app:env_protocol}.
For closed-source or API-based LLM backends whose exact parameter counts are not publicly available, we report the model identifiers, access mode, evaluation budget, and implementation settings instead.
All reported results are obtained under the fixed search budgets and evaluation protocols described in the experimental section and appendix.

\paragraph{Result reporting.}
We report correctness, Fast@1, and geometric mean speedup as the main metrics, together with level-wise KernelBench results and ablation results.
For full-suite KernelBench experiments, each task is evaluated under a fixed optimization budget, and aggregate metrics are computed over the corresponding task set.
For additional transfer and backend-sensitivity experiments, we report the task coverage and subset construction in the corresponding appendix sections.

\paragraph{Use of AI assistants.}
The authors used AI assistants only for final-stage language polishing of the manuscript.
All method design, code implementation, experimental design, experimental execution, result analysis, scientific claims, and final writing decisions were conducted, verified, and approved by the authors.

\section{Case Study: Swish Activation}
\label{app:case_study_swish}

This appendix provides the auditable trajectory for KernelBench Level-1 task 25, the Swish activation operator.
The displayed trajectory contains Steps 1--3.
The main paper reports abbreviated method names for readability, while the supplementary material provides the full executable implementations and prompts.

Starting from the PyTorch eager implementation \texttt{x * torch.sigmoid(x)}, HTAM first selects the Data Reuse direction and applies lightweight fusion to generate a scalar inline-CUDA Swish kernel.
This step avoids materializing the intermediate sigmoid output and reduces memory traffic.
HTAM then switches to Memory Access optimization, where it applies guarded \texttt{float4} vectorization and a read-only-load refinement.

\begin{figure}[H]
\centering
\includegraphics[width=1\columnwidth]{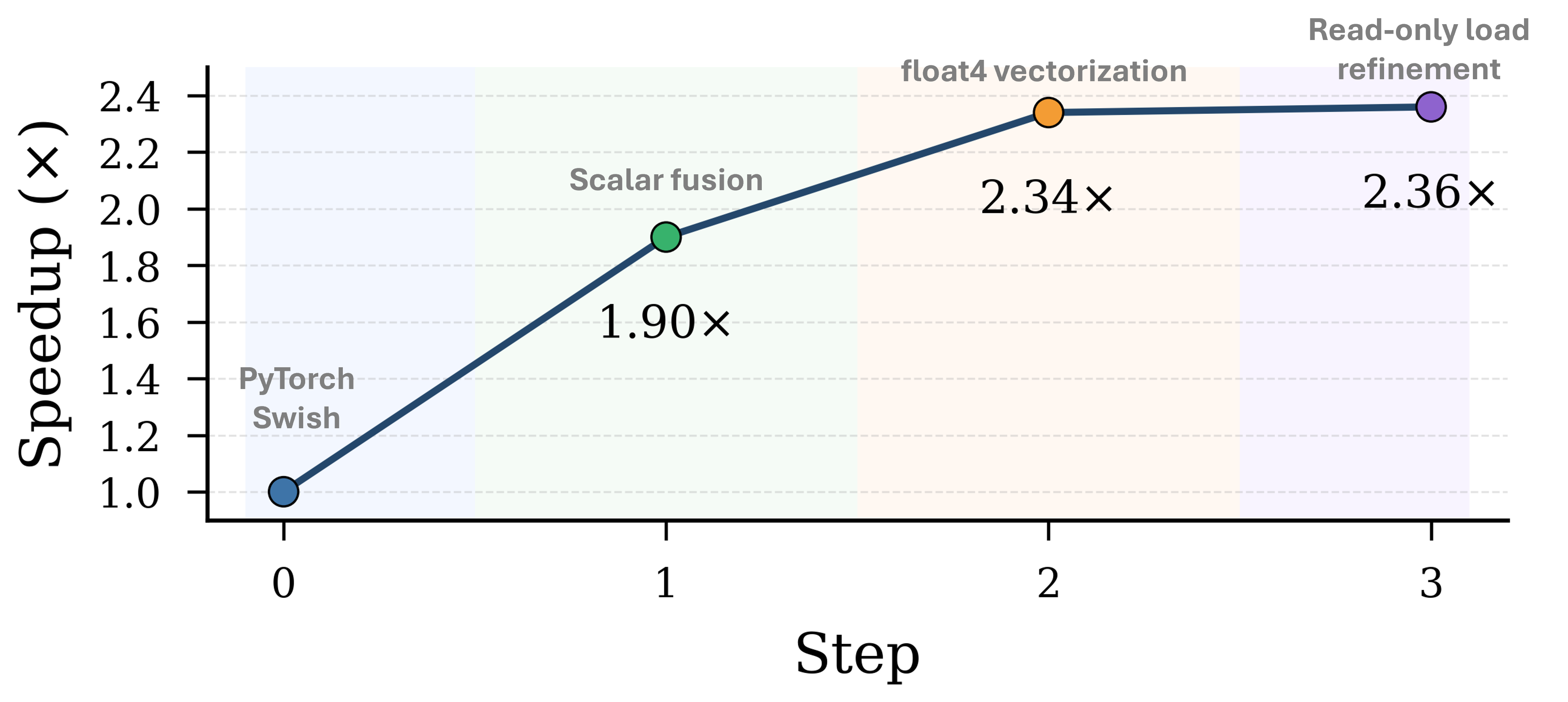}
\vspace{-0.5em}
\caption{
Step-wise speedup curve for the Swish case study.
Step 0 denotes the PyTorch eager reference.
Step 1 applies lightweight fusion under Data Reuse, while Steps 2--3 refine the fused kernel with \texttt{float4} vectorization and read-only-load optimization under Memory Access.
}
\label{fig:swish_curve_app}
\end{figure}

\subsection{Trajectory Summary}

\begin{table*}[t]
\centering
\small
\setlength{\tabcolsep}{4pt}
\renewcommand{\arraystretch}{1.08}
\begin{tabularx}{0.98\linewidth}{@{}l l l X r r@{}}
\toprule
Stage & Global direction & Local strategy & Main code change & Runtime & Speedup \\
\midrule
Baseline
& --
& --
& PyTorch Swish: \texttt{x * torch.sigmoid(x)}.
& 18.2 ms
& 1.00$\times$ \\

Step 1
& Data Reuse
& Lightweight fusion
& Fuses Swish into one scalar inline-CUDA kernel and avoids the intermediate sigmoid tensor.
& 9.56 ms
& 1.904$\times$ \\

Step 2
& Memory Access
& \texttt{float4} main loop
& Adds a vectorized main path with 16-byte alignment guards and scalar fallback.
& 7.77 ms
& 2.342$\times$ \\

Step 3
& Memory Access
& Read-only load
& Adds \texttt{\_\_ldg} and \texttt{\_\_expf} on top of the guarded vectorized path.
& 7.76 ms
& 2.345$\times$ \\
\bottomrule
\end{tabularx}
\caption{Auditable HTAM trajectory for the Swish case study. Step 1 identifies fusion as the main opportunity under Data Reuse, and Steps 2--3 further optimize memory access on the fused kernel.}
\label{tab:swish_case_trajectory}
\end{table*}

\begin{casecard}{Node mapping}
\RaggedRight
\begin{itemize}[leftmargin=1.2em, itemsep=2pt, topsep=2pt]
  \item \textbf{Step 1 global direction:}\\
  \texttt{g\_data\_reuse\_locality}\\
  Natural-language description: data reuse and locality.

  \item \textbf{Step 1 local node:}\\
  \texttt{l\_g\_reuse\_light\_epilogue\_fusion}\\
  Natural-language description: lightweight epilogue fusion for eliminating intermediate memory traffic.

  \item \textbf{Step 2 global direction:}\\
  \texttt{g\_memory\_access\_optimization}\\
  Natural-language description: memory-access optimization.

  \item \textbf{Step 2 local node:}\\
  \texttt{l\_g\_mem\_aligned\_vec4\_main\_tail}\\
  Natural-language description: \texttt{float4} vectorized main loop with alignment guards and scalar fallback.

  \item \textbf{Step 3 global direction:}\\
  \texttt{g\_memory\_access\_optimization}\\
  Natural-language description: memory-access optimization.

  \item \textbf{Step 3 local node:}\\
  \texttt{l\_g\_mem\_ldg\_readonly\_texture\_path}\\
  Natural-language description: read-only-load refinement using \texttt{\_\_ldg} and \texttt{\_\_expf}.
\end{itemize}
\end{casecard}

\subsection{Step 1: Lightweight Fusion}

\noindent\textbf{Step 1 decision and prompt excerpt.}
\begin{description}[leftmargin=0pt,itemsep=2pt,topsep=3pt]
    \item[\textbf{Input state.}]
    The original implementation is the PyTorch expression
    \texttt{x * torch.sigmoid(x)}.

    \item[\textbf{Global direction.}]
    Data reuse and locality.

    \item[\textbf{Local node.}]
    Lightweight epilogue fusion.

    \item[\textbf{Edit plan excerpt.}]
    The prompt asks the LLM to fuse the Swish computation into a single
    inline-CUDA kernel, computing
    \texttt{y[i] = x[i] / (1 + expf(-x[i]))}
    directly at writeback.
    This removes the intermediate sigmoid tensor and reduces DRAM traffic,
    while using PyTorch fallback for unsupported cases.

    \item[\textbf{Full prompt files.}]
    \texttt{step1\_local\_selection\_prompt.txt},
    \texttt{step1\_optimizer\_codegen\_prompt.txt}.
\end{description}

\noindent\textbf{Step 1 core code change.}

\noindent
The following block highlights only the main code change of Step 1.

\begin{casecodebox}{Step 1 core code change}
__global__ void swish_kernel(const float* __restrict__ x,
                             float* __restrict__ y,
                             int n) {
    int i = blockIdx.x * blockDim.x + threadIdx.x;
    int stride = blockDim.x * gridDim.x;

    for (; i < n; i += stride) {
        float xi = x[i];
        y[i] = xi / (1.0f + expf(-xi));
    }
}

torch::Tensor swish_forward(torch::Tensor x) {
    auto y = torch::empty_like(x);
    int n = x.numel();

    int threads = 256;
    int blocks = (n + threads - 1) / threads;

    swish_kernel<<<blocks, threads>>>(
        x.data_ptr<float>(),
        y.data_ptr<float>(),
        n);

    C10_CUDA_KERNEL_LAUNCH_CHECK();
    return y;
}

// Python-side dispatch keeps the original PyTorch expression as fallback.
if x.is_cuda and x.dtype == torch.float32 and x.is_contiguous():
    return ext.swish_forward(x)
return x * torch.sigmoid(x)
\end{casecodebox}

\subsection{Step 2: float4 Vectorized Main Loop}

\noindent\textbf{Step 2 decision and prompt excerpt.}
\begin{description}[leftmargin=0pt,itemsep=2pt,topsep=3pt]
    \item[\textbf{Input state.}]
    The current implementation is a fused scalar CUDA Swish kernel.

    \item[\textbf{Global direction.}]
    Memory-access optimization.

    \item[\textbf{Local strategy.}]
    \texttt{float4} vectorized memory access.

    \item[\textbf{Edit plan excerpt.}]
    The prompt asks the LLM to use a \texttt{float4} main path when input
    and output pointers are 16-byte aligned, require \texttt{n \% 4 == 0}
    before reinterpretation, and preserve a scalar fallback for all other cases.

    \item[\textbf{Full prompt file.}]
    \texttt{step2\_local\_selection\_prompt.txt},
    \texttt{step2\_optimizer\_codegen\_prompt.txt}.
\end{description}

\noindent\textbf{Step 2 core code change.}

\noindent
The following block highlights only the main code change of Step 2.

\begin{casecodebox}{Step 2 core code change}
__device__ __forceinline__ float _swish(float v) {
    return v / (1.0f + expf(-v));
}

__global__ void swish_vec4_kernel(const float4* __restrict__ x4,
                                  float4* __restrict__ y4,
                                  int n4) {
    int i = blockIdx.x * blockDim.x + threadIdx.x;
    int stride = blockDim.x * gridDim.x;

    for (; i < n4; i += stride) {
        float4 v = x4[i];
        v.x = _swish(v.x);
        v.y = _swish(v.y);
        v.z = _swish(v.z);
        v.w = _swish(v.w);
        y4[i] = v;
    }
}

const float* xptr = x.data_ptr<float>();
float* yptr = y.data_ptr<float>();

const bool aligned_in  = (reinterpret_cast<uintptr_t>(xptr) % 16 == 0);
const bool aligned_out = (reinterpret_cast<uintptr_t>(yptr) % 16 == 0);
const bool n_mod4_ok   = ((n % 4) == 0);

if (aligned_in && aligned_out && n_mod4_ok) {
    const int n4 = n / 4;
    int blocks = (n4 + threads - 1) / threads;
    if (blocks > 65535) blocks = 65535;

    swish_vec4_kernel<<<blocks, threads>>>(
        reinterpret_cast<const float4*>(xptr),
        reinterpret_cast<float4*>(yptr),
        n4);
} else {
    int blocks = (n + threads - 1) / threads;
    if (blocks > 65535) blocks = 65535;

    swish_scalar_kernel<<<blocks, threads>>>(xptr, yptr, n);
}
\end{casecodebox}

\subsection{Step 3: Read-only-load Refinement}

\noindent\textbf{Step 3 decision and prompt excerpt.}
\begin{description}[leftmargin=0pt,itemsep=2pt,topsep=3pt]
    \item[\textbf{Input state.}]
    The current implementation already contains a \texttt{float4} vectorized Swish kernel.

    \item[\textbf{Global direction.}]
    Memory-access optimization.

    \item[\textbf{Local strategy.}]
    Read-only-load refinement.

    \item[\textbf{Edit plan excerpt.}]
    The prompt asks the LLM to use \texttt{\_\_ldg} for read-only input loads,
    apply this path to both vectorized and scalar kernels, use \texttt{\_\_expf}
    for the sigmoid term, and keep output stores unchanged.

    \item[\textbf{Full prompt file.}]
    \texttt{step3\_local\_selection\_prompt.txt},
    \texttt{step3\_optimizer\_codegen\_prompt.txt}.
\end{description}

\noindent\textbf{Step 3 core code change.}

\noindent
The following block highlights only the main code change of Step 3.

\begin{casecodebox}{Step 3 core code change}
__device__ __forceinline__ float _swish_fast(float v) {
    return v / (1.0f + __expf(-v));
}

__global__ void swish_vec4_ldg_kernel(const float4* __restrict__ x4,
                                      float4* __restrict__ y4,
                                      int n4) {
    int i = blockIdx.x * blockDim.x + threadIdx.x;
    int stride = blockDim.x * gridDim.x;

    for (; i < n4; i += stride) {
        float4 v = __ldg(x4 + i);
        v.x = _swish_fast(v.x);
        v.y = _swish_fast(v.y);
        v.z = _swish_fast(v.z);
        v.w = _swish_fast(v.w);
        y4[i] = v;
    }
}

__global__ void swish_scalar_ldg_kernel(const float* __restrict__ x,
                                        float* __restrict__ y,
                                        int n) {
    int i = blockIdx.x * blockDim.x + threadIdx.x;
    int stride = blockDim.x * gridDim.x;

    for (; i < n; i += stride) {
        float xi = __ldg(x + i);
        y[i] = xi / (1.0f + __expf(-xi));
    }
}

const int threads = 512;

if (aligned_in && aligned_out && n_mod4_ok) {
    const int n4 = n / 4;
    int blocks = (n4 + threads - 1) / threads;
    if (blocks > 65535) blocks = 65535;

    swish_vec4_ldg_kernel<<<blocks, threads>>>(
        reinterpret_cast<const float4*>(xptr),
        reinterpret_cast<float4*>(yptr),
        n4);
} else {
    int blocks = (n + threads - 1) / threads;
    if (blocks > 65535) blocks = 65535;

    swish_scalar_ldg_kernel<<<blocks, threads>>>(xptr, yptr, n);
}
\end{casecodebox}

\subsection{Case-study takeaway.}
This case study highlights three practical properties of HTAM. 
First, the optimization trajectory is not a black-box sequence of unrelated trials: HTAM first selects a data-reuse direction to fuse the Swish computation, and then shifts to memory-access optimization for guarded vectorization and read-only-load refinement. 
Second, each step corresponds to an executable inline-CUDA edit rather than a natural-language hint alone. 
The generated kernels include concrete implementation details such as grid-stride loops, alignment checks, scalar fallbacks, \texttt{float4} reinterpretation, and read-only loads, making the trajectory inspectable at the code level. 
Third, the final improvement is accumulated through small, safe transformations instead of a single aggressive rewrite. 
This behavior is important for operator optimization, where correctness constraints and boundary cases often make direct large edits brittle. 
Overall, the Swish trajectory illustrates how hierarchical transition memory can first guide a high-level optimization shift and then refine the selected direction into executable CUDA implementations.

The full executable implementations and code-generation prompts for all three steps are provided in the supplementary material:
\path{supplementary/case_study/swish/}.

This directory contains one subdirectory for each step, including the generated Python implementation and the corresponding code-generation prompt. 
Each implementation includes the \texttt{load\_inline} wrapper, generated CUDA/C++ kernel, \texttt{ModelNew} class, input helpers, and PyTorch fallback path.

\end{document}